\documentclass[3p,twocolumn, round, sort & compress]{elsarticle}

\usepackage{array}
\usepackage{hyperref}
\usepackage{graphicx}
\usepackage{times}
\usepackage{epsfig}
\usepackage{graphicx}
\usepackage{graphicx}
\usepackage{amsmath,cases,amssymb}
\usepackage{mathrsfs}
\usepackage{amsmath,bm}
\usepackage{algorithm}
\usepackage{caption}
\usepackage{algorithmic}
\usepackage{bm}
\usepackage{multirow}
\usepackage{mathrsfs}
\usepackage{wrapfig}
\usepackage{booktabs}
\usepackage{color}
\usepackage{verbatim} %comments
\usepackage[switch]{lineno}

\usepackage{url}
\modulolinenumbers[1]

\newif\ifshowfig\showfigtrue

\biboptions{authoryear}

\begin{document}

\begin{frontmatter}

\title{An Efficient Multitask Neural Network for Face Alignment, Head Pose Estimation and Face Tracking}

\author{Jiahao Xia }
\ead{Jiahao.Xia@student.uts.edu.au}

\author{Haimin Zhang}
\ead{Haimin.Zhang@uts.edu.au}

\author{Shiping Wen}
\ead{Shiping.Wen@uts.edu.au}

\author{Shuo Yang}
\ead{Shuo.Yang@student.uts.edu.au}

\author{ Min Xu \corref{mycorrespondingauthor}}
\cortext[mycorrespondingauthor]{Corresponding author}

\ead{Min.Xu@uts.edu.au}

\address{School of Electrical and Data Engineering, Faculty of Engineering and IT, University of Technology Sydney }
\address{15 Broadway, Ultimo, NSW 2007, Australia}

\begin{abstract}
While Convolutional Neural Networks (CNNs) have significantly boosted the performance of face related algorithms, maintaining accuracy and efficiency simultaneously in practical use remains challenging. The state-of-the-art methods employ deeper networks for better performance, which makes it less practical for mobile applications because of more parameters and higher computational complexity. Therefore, we propose an efficient multitask neural network, Alignment \& Tracking \& Pose Network (ATPN) for face alignment, face tracking and head pose estimation. Specifically, to achieve better performance with fewer layers for face alignment, we introduce a shortcut connection between shallow-layer and deep-layer features. We find the shallow-layer features are highly correspond to facial boundaries that can provide the structural information of face and it is crucial for face alignment. Moreover, we generate a cheap heatmap based on the face alignment result and fuse it with features to improve the performance of the other two tasks. Based on the heatmap, the network can utilize both geometric information of landmarks and appearance information for head pose estimation. The heatmap also provides attention clues for face tracking. The face tracking task also saves us the face detection procedure for each frame, which also significantly boost the real-time capability for video-based tasks. We experimentally validate ATPN on four benchmark datasets, WFLW, 300VW, WIDER Face and 300W-LP. The experimental results demonstrate that it achieves better performance with much less parameters and lower computational complexity compared to other light models.
%While convolutional neural networks (CNNs) have significantly boosted the performance of face related algorithms, maintaining accuracy and efficiency simultaneously in practical use remains challenging. In this paper, we present a multitask CNN, ``Alignment \& Pose \& Tracking Network (ATPN)", which can simultaneously achieve face alignment, head pose estimation and face tracking with excellent real-time speed and high accuracy in videos. Concretely, in order to localize facial landmark accurately with fewer layers, we propose a novel architecture which utilizes the structural information contained in low-level features for face alignment. Besides, the conception of coordinates is introduced into CNN by adopting CoordConv in the last few layers to improve the accuracy further without sacrificing efficiency. Then, a cheap heatmap is generated directly based on the predicted facial landmark to introduce an attention mechanism into CNN, by which the model can take advantages of both geometric and appearance information to estimate head pose. Moreover, the face tracking task makes face detection unnecessary in each frame to increase the speed of video-based processing drastically. The proposed framework is evaluated on four benchmark datasets, WFLW, 300VW, WIDER Face and 300W-LP.  The experimental results show that the ATPN achieves better performance than previous state-of-the-art methods with less number of parameters and FLOPS.
\end{abstract}


\end{frontmatter}

\section{Introduction} \label{sec:intro}

%and other prediction tasks \citep{MEP, PPO, FOG}
Face alignment and head pose estimation, which aims at localizing a group of predefined facial landmarks and estimating the Euler angle of head, have a wide range of applications, such as facial expression recognition \citep{facial_expression}, face morphing \citep{face_morphing} and 3D face reconstruction \citep{face_recostruction}. Although CNNs have boosted the performance of face-related problems and other prediction tasks significantly \citep{ANNA, ANNW, MEP, PPO, FOG}, the low real-time capability is still the main barrier to practical use since many applications run on the mobile devices \citep{fatigue}. Developing a practical face alignment and head pose estimation method remains challenging.
% However, the biggest barrier between algorithms and practical use is that most algorithms cannot obtain real-time performance and high accuracy at the same time. Therefore, developing a practical face alignment and head pose estimation method remains challenging.

The facial structural information contained in facial boundaries is crucial to face alignment \citep{PropNet}, especially in unconstrained scenarios (large pose, extreme lighting condition and occlusion). To explicitly utilize the structural information, many recent studies \citep{LAB, PropNet} utilize several cascaded hourglass modules \citep{Hourglass} to generate facial boundary heatmaps and the heatmaps are then fused with intermediate features. However, compared to regressing landmarks from the input image directly, these methods are much less efficient because of a large number of bottom-up and top-down convolution layers.

\begin{figure}[t!]
	\centering
	\includegraphics[width=\linewidth]{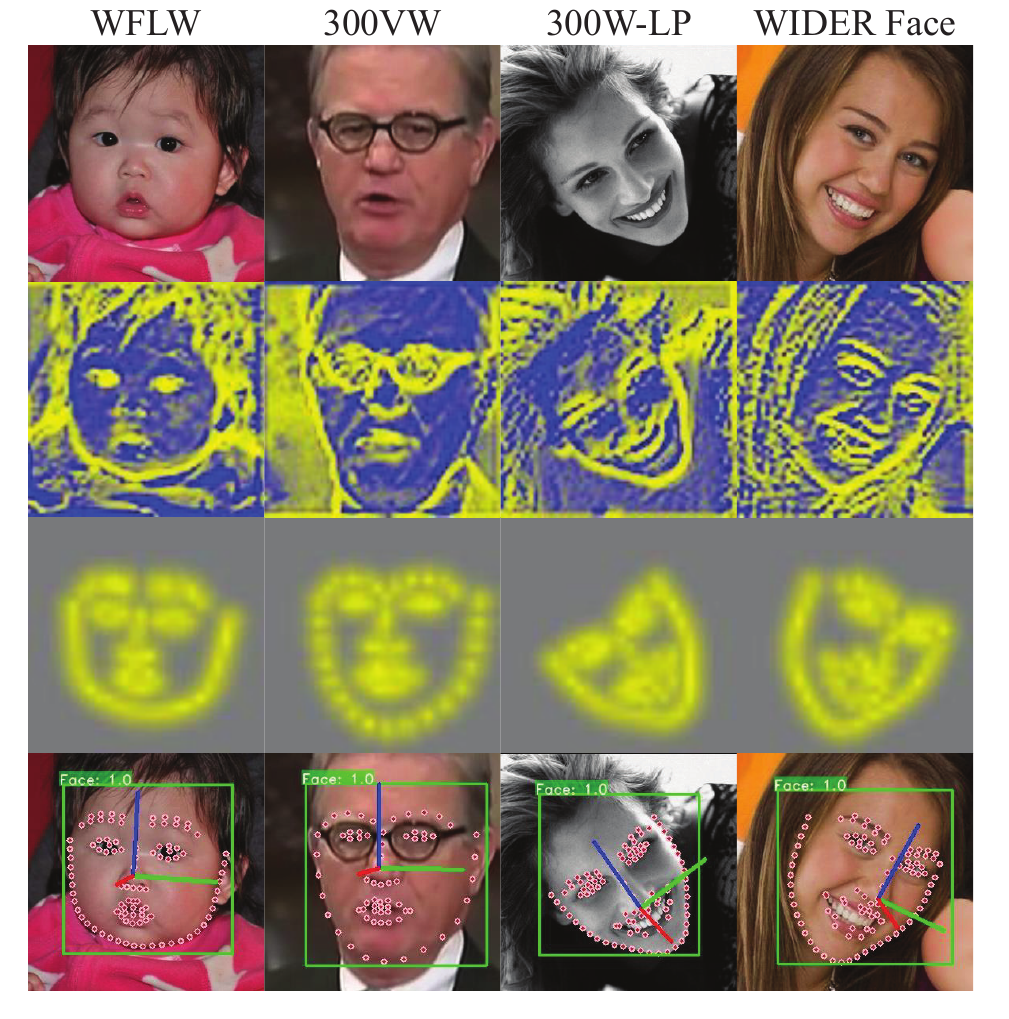}
	\caption{The first row shows the input images from different datasets and the second row illustrates the features maps at low-level. The heatmaps generated by face alignment result, shown in the third row, can provide attention clues for the CNN. The fourth row shows the estimated results of the ATPN (the {\color{red} \textbf{red axis}} points towards the front of the face, {\color{blue} \textbf{blue}} pointing upward and {\color{green} \textbf{green}} pointing left side).} 
	\label{fig1}
\end{figure}

Existing head pose estimation methods only utilize the geometric information of facial landmarks (model-based approach \citep{accurate_model,Openface}) or appearance information of the input image (appearance-based approach \citep{Robust_pose,Ruzi, Multitask-pose, Hyperface}) to estimate the Euler angle of faces. Model-based approaches fit a predefined 3D face model to the face in image according to facial landmarks. As a result, they ignore the appearance information of image features. Appearance-based approaches only rely on the appearance information of the input image. Nevertheless, the features outputted by a CNN or manifold embedding do not necessarily correspond to pose angle \citep{Robust_pose}, which leads to fragile robustness in unconstrained scenarios.  

Moreover, most existing face alignment and head pose estimation algorithms require a face detector before the process. Therefore, they cannot run as fast as the speed in theory. In the video-based processing, face tracking makes face detection unnecessary in each frame, by which face alignment and head pose estimation can be accelerated further.

In this paper, we find out that the shallow-layer features are highly correspond to facial boundaries and they contain the facial structural information(Fig.1 $2th$ column). We give a shortcut to the features in shallow layers so that a light model can also explicitly employ the structural information for accurate face alignment. Then, a cheap heatmap (Fig.1 $3rd$ column) is generated directly based on face alignment result and fused with the intermediate features. Based on the procedure, the ATPN can utilize both the appearance information of input image and the geometric information of the facial landmarks to achieve accurate and robust head pose estimation. It also provides the attention clues for face tracking task. Moreover, the face tracking task can save the face detection procedure to further accelerated face alignment and head pose estimation in video-based processing. 
%makes the ATPN more practical because face detection is unnecessary in each frame of video-based processing.

In a nutshell, our main contributions include:
\begin{itemize} 
	\item Proposing a light architecture that employs the structural information in low-level features for the face alignment. Compared to other light models, it achieves the best performance with the least parameters and lowest computational complexity.
	\item Providing the geometric information for head pose estimation and attention clues for face tracking by a heatmap generated from the face alignment results. 
	\item Proposing a practical multitask framework for face alignment, head pose estimation and face tracking for video-based processing.
	\item Conducting extensive experiments and ablation studies on various datasets to prove the effectiveness of ATPN.
\end{itemize}

\section{Related Work}
\subsection{Face Alignment}
In the early stage, face alignment algorithms can be split into two categories: the model-based method \citep{CE-CLM} and the regression-based method \citep{ESR, SDM, CFSS, ACFSS}. With the development of neural network, the performance of face alignment has been improved significantly. Many frameworks, such as DVLN \citep{DVLN} and UFLD \citep{TUF}, achieve higher accuracy and stronger robustness by taking advantages of the stronger expressiveness of CNNs. To improve the performance of face alignment further, \citep{SAN, AVS, 3FabRec, DeCaFA, SBR} utilize style transfer or semi-supervised learning to augment training datasets, which addresses the problem of scarce training samples. However, with more large-scale datasets \citep{LAB, LAPA} released, the number of training samples is not the main issue that limits the performance of face alignment. \citep{LAB} find the structural information contained in facial boundary is crucial to face alignment. Following their work, \citep{PropNet} present a Landmark-Boundary Propagation Module to generate more accurate boundary heatmaps; \citep{ADC} utilize regional latent heatmap to achieve attention-driven face alignment in very high resolution. Nevertheless, it is difficult for the light model \citep{KWTQ, RFA} to utilize the information since generating an accurate boundary heatmap requires a very deep network. Therefore, maintaining both real-time capability and performance is quite challenging to face alignment.

In terms of video-based face alignment, many works try to model the temporary dependency explicitly for better performance. \citep{Two-Stream-action} propose a two-stream architecture to utilize single frame and the optical flow of multi-frame simultaneously; \citep{Two-stream_video} optimize the method further by encoding the video input as active appearance codes in temporal stream. To improve the performance in motion-blurred video, \citep{FAB} deblur the image based on two previous boundary maps and recent frames. However, these methods are also too complex to run on the mobile devices. 

\subsection{Head Pose Estimation}
\citep{U3D-pose} has proven that the pose of a 3-point configuration is uniquely determined up to a reflection by its 2D projection under weak perspective model. Based on the conclusion, many works, such as \citep{accurate_model}, fit a predefined 3D face model to images based on the geometric information of facial landmark to estimate Euler angles, which is named as the model-based method. To address the problem that model-based methods degrade dramatically in large pose, \citep{Openface} replace the 2D landmarks by the 3D landmarks \citep{CE-CLM} that contains more geometric information. However, compared to the appearance information, the geometric information of 3D landmarks is still insufficient for robust head pose estimation.
%utilize CE-CLM\citep{CE-CLM} to localize the 3D position of facial landmark and then fit a model which is generated by training dataset to image to improve the accuracy of head pose estimation.

Different from model-based methods, appearance-based methods regress head pose directly from images. \citep{Robust_pose} adopt a mixture of linear regressions with partially-latent output to estimate head pose; \citep{CNN-Pose} utilize CNN to regress head pose and study the influence of different optimizers and CNN structures; \citep{Ruzi} and \citep{FSA-Net} classify head pose into several ranges and then regress a fine-grained result based on the range. Nevertheless, the features outputted by a CNN or manifold embedding do not necessarily correspond to pose angle because of the lack of attention clues.

\begin{figure*}[t!]
	\centering
	\includegraphics[width=\linewidth]{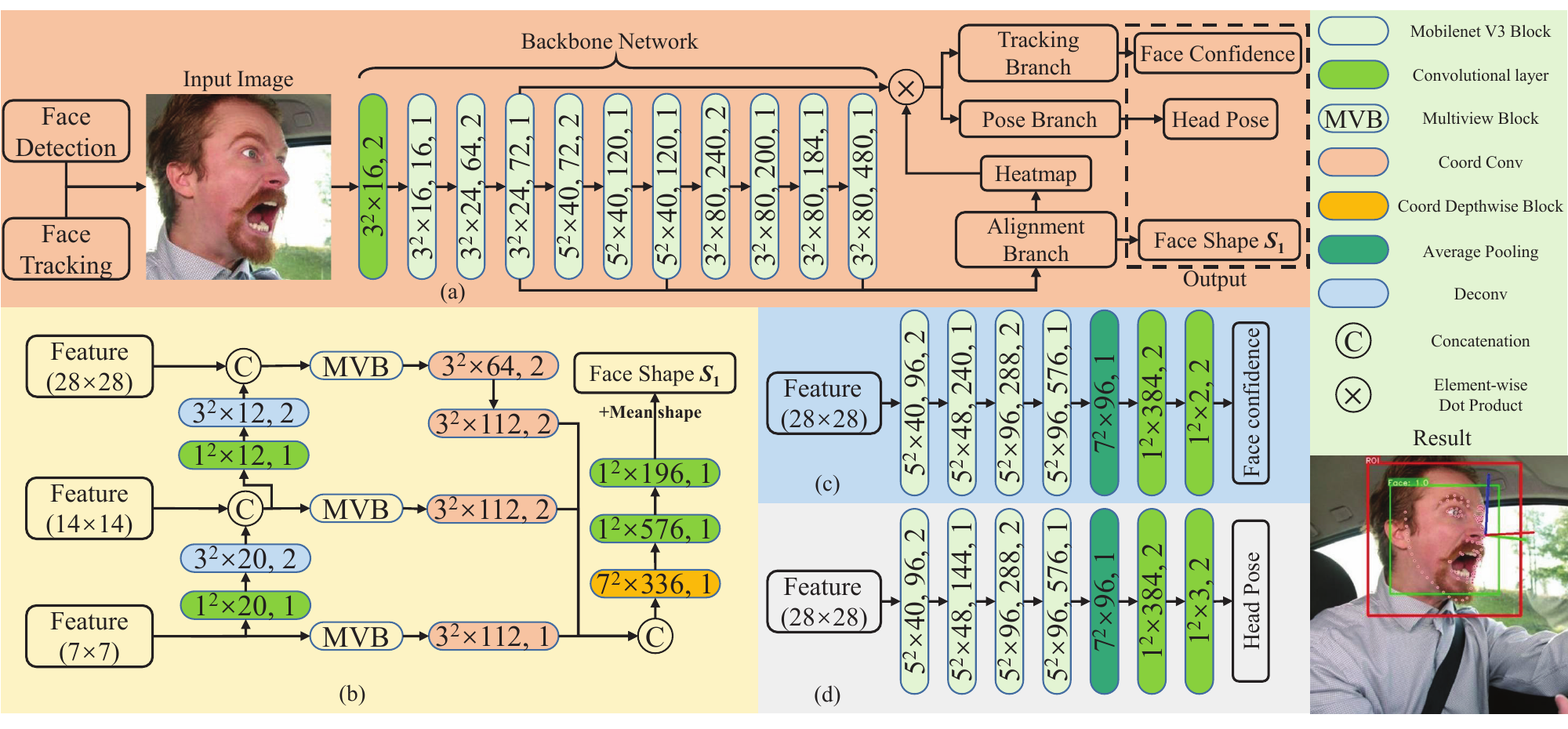}
	\caption{An overview of our multitask framework. The kernels of Mobilenet V3 block are described as kernel size $\times$ channels, expansion size, stride. And the kernels in other layers are described as kernel size $\times$ channels, stride. (a) The overall framework. The backbone architecture based on Moblienet-V3 module is used to extract image features for facial landmarks localization, head pose estimation and face tracking. (b) The alignment branch is used for facial landmarks localization with multiscale features. (c) The tracking branch output a confidence of face based on the features fused with heatmap. (d) The pose branch regresses head pose from the features fused with heatmap.}
	\label{fig2}
\end{figure*}

\subsection{Multitask Network}
Multitask learning enables CNNs to learn a task together with other related tasks simultaneously using a shared representation. It improves the performance and real-time speed significantly. \citep{zhang-multi} propose a one-stage multitask CNN to three face-related tasks: face detection, face view prediction and 5 facial landmarks localization. \citep{MTCNN} optimize the framework further by a cascaded network; \citep{Multitask-pose} propose a cascaded framework for face detection and head pose estimation; \citep{Hard-MTCNN} present a hardwired accelerator for the framework. Hyperface \citep{Hyperface} utilize both the global and local information on face for four tasks (face detection, landmarks localization, head pose estimation and gender recognition). Nevertheless, multitask framework can share more information for better performance, such as geometric information. Unfortunately, there is no existing work yet.

\section{Approach}
In this section, we introduce a multitask framework, ATPN, for face alignment, head pose estimation and face tracking. As illustrated on Fig.2(a), the ATPN consists of four parts: the backbone network, alignment branch, pose branch and tracking branch. The input is the Region of Interest (ROI) of face that is acquired by a face detector or the minimum bounding rectangle of the face landmarks in previous frames (25\% extension of each boundary). The backbone network firstly learns features from input image with Mobilenet-V3 block \citep{mobileV3}. Then the alignment branch localize facial landmarks by the features at different levels (28$\times$28, 14$\times$14, 7$\times$7). Based on the predicted facial landmarks, a heatmap is generated directly and fused with low-level features (28$\times$28) to provide geometric information and attention clue for other tasks. Finally, the two branch regress the Euler angles and the confidence of face respectively. If the confidence is larger than a certain threshold (0.7 for ATPN), the ROI of next framework is calculated based on the predicted facial landmarks. 

\subsection{Alignment Branch}
The structural information of the facial boundaries is crucial for facial alignment. Different from other works \citep{LAB, PropNet} that utilize an additional network or branch to generate boundary heatmaps, we directly employ the structural information in low-level features, as shown in Fig.2(b). Referred to feature pyramid network \citep{FPN}, we firstly fuse the low-level features with high-level features by $1\times1$ Conv and Deconv layers. Then, the features at different levels are fed into the multiview block \citep{MVB}. As illustrated in Fig.3, it consists of three CNN layers to create three different receptive fields. Besides, the residual connection of the multiview block also preserves the low-level features. The outputs of the multiview block are downsampled directly by CoordConv \citep{CoordCNN}. On the one hand, it shorten the information path between low-level featuress and output layers. On the other hand, it introduces a coordinate conception into CNN, which is significant to coordinate regression. Finally, we utilize a $7 \times 7$ Coord Depthwise Block to project the feature map into a vector. Compared to pooling or linear layers, the Coord Depthwise Block can preserve the spatial structure of the input image. Instead of predicting the facial landmarks directly, the alignment branch predicts the residual error between target face shape $\bm{S_1}$ and mean shape $\bm{S_0}$. The mean shape $\bm{S_0}$ provides a good initial face shape for the framework to make the result more stable.    
 
\begin{figure}[t!]
	\centering
	\includegraphics[width=\linewidth]{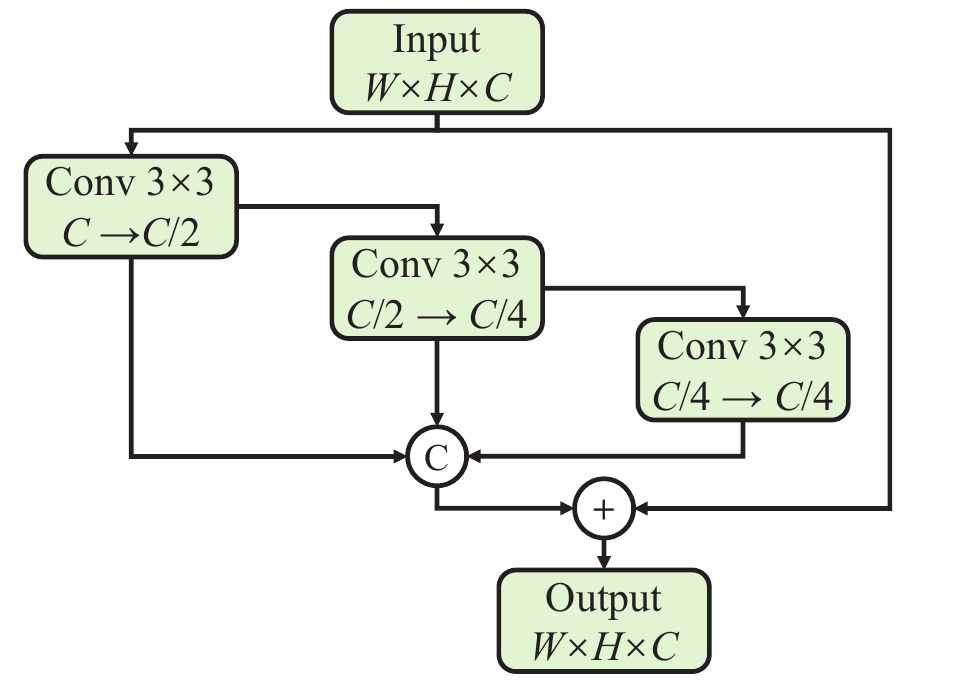}
	\caption{The structure of a multiview block. \emph{W}, \emph{H}, \emph{C} indicate the width, hight and channel of feature respectively}
	\label{fig3}
\end{figure}

\subsection{Pose Branch and Tracking Branch}
The tracking branch (Fig.2(c)) and pose branch (Fig.2(d)) mainly consist of MobileNet-V3 Block for better efficiency. In the pose branch, the output layer is activated by Tanh and then multiplied by $\pi$ to normalize the output $\bm{P_h}$ into $[-\pi, +\pi]$. Hence, the predicted result can be more stable. In the tracking branch, the output is activated by Softmax to normalize face confidence $C_f$ in [0, 1].

\subsection{Heatmap}
Different from the heatmap in other works \citep{LAB, DeCaFA, PropNet}, the heatmap in ATPN is generated directly based on the predicted landmarks by a formula. Therefore, the parameters and computational complexity can be reduced significantly. The formula can be written as:
\begin{equation}
	H\left( {x,y} \right) = \frac{1}{{\sqrt {1 + {{\min }_{({x_i^\prime},{y_i^\prime}) \in {{\bm{S}}_{1}}}}\left\| {\left( {x,y} \right) - \left( {{x_i^\prime},{y_i^\prime}} \right)} \right\|} }},
\end{equation}
where $H(x,y)$ is the intensity of point $(x,y)$. $(x_i^\prime,y_i^\prime)$ indicates the coordinate for the $i$-th landmark of $\bm{S}_1$. To avoid the problem that the CNN completely ignore the feature far from facial landmarks, $H(x,y)$ is set to 0.5 if the value is less than 0.5. Then, heatmap is fused with features by element wise multiplication, as shown below:
\begin{equation}
{\bm{F}_O} = \bm{F}_I \otimes \bm{H},
\end{equation}
where $\bm{F}_O$ indicates the output features, $\bm{H}$ and $\bm{F_I}$ indicates the heatmap and the features learned by backbone network. By fusing the heatmap with the intermediate features, the output features contains both appearance information of the input images and the geometric information of facial landmarks. Moreover, the heatmap also provides the attention clue for ATPN to eliminate the interference of background and improve the performance of face tracking.

%As a result, the features near facial boundary are highlight and interference of background is restrained. Moreover, pose branch can utilize the geometric information contained in heatmap and appearance information contained in input feature to estimate the Euler angle of head.

\subsection{Training Strategy}
We train the ATPN with three stages to alleviate the problem that a difficult task is still underfitting while a easy task is overfitting. 

%In experiment, we find the difficulty of the three tasks is quite various. Therefore, it is quite common that the face tracking task has already been overfitting while the face alignment task is still underfitting. In order to address this problem, the ATPN is trained using three stages and the framework achieve three tasks simultaneously in validation.

In the first stage, we train the backbone network and face alignment branch together. We calculate the point-to-point Euclidean and normalize it with the Inter-ocular distance \citep{300W}, which can be written as:
\begin{equation}
{L_A} = \frac{\left\|\bm{S}_1-\bm{{S}}_1^\prime\right\|_2}{\bm{d}_{ION}},
\end{equation}
where $L_A$ indicates the normlizaed error of face alignment. $\bm{d}_{ION}$ is the Inter-ocular distance, $\bm{{S}}_1^\prime$ and $\bm{{S}}_1$ are the predicted and annotated landmarks respectively. Then, the loss function is formulated as:
\begin{equation}
\min \left( {\frac{1}{{{N_1}}}\left( {\sum\limits_{i = 1}^{{N_1}} {{L_{A - i}}} } \right) + {w_1}{l_{2 - A}}} \right),
\end{equation}
where $N_1$ indicates the number of samples which are used in stage 1. $l_{2-A}$ is the L2-regularization loss of backbone network and face alignment branch, and $w_1$ is the weight of $l_{2-A}$.

\begin{figure}[t!]
	\centering
	\includegraphics[width=\linewidth]{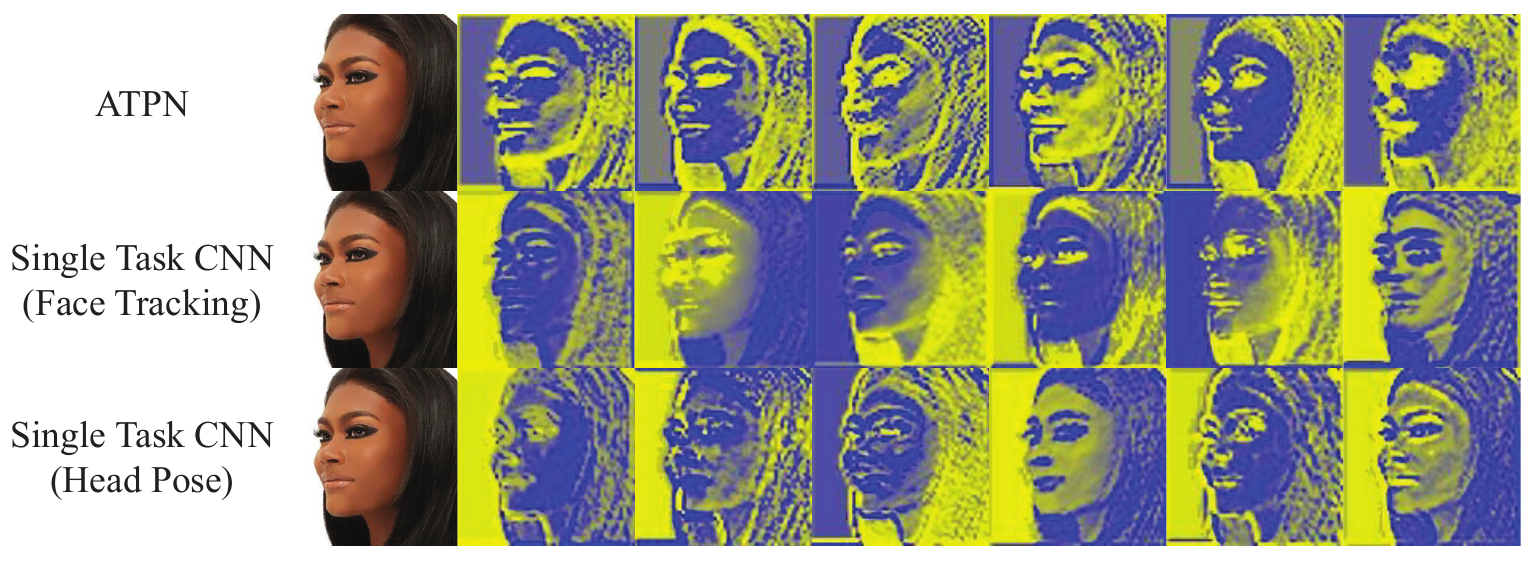}
	\caption{Column 1 shows the input image and column 2 to column 7 illustrate the low-level feature maps in the ATPN and two single task CNN.}
	\label{fig4}
\end{figure}

Then, we freeze the weights of the backbone and train the tracking branch and pose branch in stage 2 and stage 3 respectively. The face tracking error $L_T$ can be written as:
\begin{equation}
{L_T} =  - \left( {{y_t}\log \left( {{p_t}} \right) + \left( {1 - {y_t}} \right)\left( {1 - \log \left( {{p_t}} \right)} \right)} \right),
\end{equation}
where $p_t$ is the predicted face confidence and $y_i\in[0,1]$ is the annotation (0: background, 1: face). The learning object of the second stage can be written as:
\begin{equation}
\min \left( {\frac{1}{{{N_2}}}\left( {\sum\limits_{i = 1}^{{N_2}} {{L_{T-i}}} } \right){\rm{ + }}{w_2}{l_{2 - t}}} \right),
\end{equation}
where $N_2$ indicates the number of samples used in the second stage. $l_{2-t}$ is the L2-regularization loss of the tracking branch and $w_2$ is the weight of $l_{2-f}$. The error of head pose estimation $L_p$ can be written as:
\begin{equation}
{L_P} = \frac{{\sqrt {\left\| {{{\bf{P}}_h} - {{\bf{P}}_h}^\prime } \right\|} }}{3},
\end{equation}
where $\bm{P}_h$ and $\bm{P}_h^\prime$ are the predicted and annotated head pose respectively. Finally, the learning object of third stage can be written as:
\begin{equation}
\min \left( {\frac{1}{{{N_3}}}\left( {\sum\limits_{i = 1}^{{N_3}} {{L_{P-i}}} } \right){\rm{ + }}{w_3}{l_{2 - p}}} \right),
\end{equation}
where $N_3$ indicates the number of samples used in the third stage. $l_{2-p}$ is the L2-regularization loss of pose branch and $w_3$ is the weight of $l_{2-p}$.

\begin{table*}[t!]
	\centering
	\begin{tabular}{m{1.4cm}<{\centering}|m{3.8cm}<{\centering}|m{0.8cm}<{\centering}|m{0.7cm}<{\centering}|m{1.4cm}<{\centering}|m{1.5cm}<{\centering}|m{1.3cm}<{\centering}|m{1.2cm}<{\centering}|m{0.7cm}<{\centering}}
		\hline
		Metric & Method & Testset & Pose  & Expression & Illumination & Make-up & Occlusion & Blur \\ \hline
		\multirow{18}{*}{NME(\%)$\downarrow$} 
		& LAB$^\dag$ \citep{LAB} & 5.27 & 10.24 & 5.51 & 5.23 & 5.15 & 6.79  & 6.32  \\
		& SAN$^\dag$ \citep{SAN} & 5.22 & 10.39 & 5.71 & 5.19 & 5.49 & 6.83 & 5.8  \\
		& 2 Hourglass$^\dag$ \citep{Hourglass} & 5.19 & 9.03 & 5.53 & 5.08 & 4.97 & 6.45 & 5.93 \\
		& DeCaFA$^\dag$ \citep{DeCaFA} & {\color{blue} \textbf{4.62}} & {\color{blue} \textbf{8.11}} & {\color{blue} \textbf{4.65}} & {\color{blue} \textbf{4.41}} & {\color{blue} \textbf{4.63}} & {\color{blue} \textbf{5.74}} & {\color{blue} \textbf{5.38}}  \\
		& PropNet$^\dag$ \citep{PropNet}  & {\color{red} \textbf{4.05}} & {\color{red} \textbf{6.92}} & {\color{red} \textbf{3.87}} & {\color{red} \textbf{4.07}} & {\color{red} \textbf{3.76}} & {\color{red} \textbf{4.58}} & {\color{red} \textbf{4.36}} \\ \cline{2-9}
		& MuSiCa98$^\ddag$ \citep{KWTQ} & 7.90 & 15.80 & 8.52 & 7.49 & 8.56 & 10.04 & 8.92 \\
		& 3FabRec$^\ddag$ \citep{3FabRec} & 5.62 & 10.23 & 6.09 & 5.55 & 5.68 & {\color{red} \textbf{5.92}} & 6.38 \\
		& $G\&LSR_\omega$$^\ddag$\citep{RFA} & 5.26 & - & - & - & - & - & -\\
		& Res18+AVS$^\ddag$ \citep{AVS} & {\color{blue} \textbf{5.25}} & {\color{blue} \textbf{9.10}} & {\color{blue} \textbf{5.83}} & {\color{red} \textbf{4.93}} & {\color{blue} \textbf{5.47}} & {\color{blue} \textbf{6.26}} & {\color{blue} \textbf{5.86}} \\
		& ATPN & {\color{red} \textbf{5.13}} & {\color{red} \textbf{8.97}} & {\color{red} \textbf{5.49}} & {\color{blue} \textbf{4.95}} & {\color{red} \textbf{4.94}} & 6.30 & {\color{red} \textbf{5.78}} \\ \hline
		\multirow{10}{*}{FR$_{0.1}$(\%)$\downarrow$} 
		& LAB$^\dag$ & 7.56 & 28.83 & 6.37 & 6.73 & 7.77 & 13.72 & 10.74 \\
		& SAN$^\dag$ & 6.32 & 27.91 & 7.01 & 4.87 & 6.31 & 11.28 & {\color{blue} \textbf{6.60}}   \\
		& 2 Hourglass$^\dag$ & 6.04 & 25.46 & 5.41 & 5.59 & {\color{blue} \textbf{5.34}} & 12.5 & 8.40   \\
		& DeCaFA$^\dag$ & {\color{blue} \textbf{4.84}} & {\color{blue} \textbf{21.4}} & {\color{blue} \textbf{3.73}} & {\color{blue} \textbf{3.22}} & 6.15 & {\color{blue} \textbf{9.26}} & 6.61  \\
		& PropNet$^\dag$ & {\color{red} \textbf{2.96}} & {\color{red} \textbf{12.58}} & {\color{red} \textbf{2.55}} & {\color{red} \textbf{2.44}} & {\color{red} \textbf{1.46}} & {\color{red} \textbf{5.16}} & {\color{red} \textbf{3.75}}  \\ 
		\cline{2-9}
		& MuSiCa98$^\ddag$ & - & - & - & - & - & - & - \\
		& 3FabRec$^\ddag$   & 8.28    & 34.35 & {\color{blue} \textbf{8.28}}       & 6.73         & 10.19   & 15.08     & 9.44  \\
		& $G\&LSR_\omega$$^\ddag$ & {\color{red} \textbf{5.72}} & - & - & - & - & - & -\\
		& Res18+AVS$^\ddag$ & 7.44    & {\color{blue} \textbf{32.52}} & 8.3 & {\color{red} \textbf{4.3}} & {\color{blue} \textbf{8.25}} & {\color{blue} \textbf{12.77}} & {\color{blue} \textbf{9.06}}  \\
		& ATPN & {\color{blue} \textbf{6.27}} & {\color{red} \textbf{26.99}} & {\color{red} \textbf{6.05}} & {\color{blue} \textbf{4.72}} & {\color{red} \textbf{7.28}} & {\color{red} \textbf{12.22}} & {\color{red} \textbf{7.89}} \\ \hline
		\multirow{10}{*}{AUC$_{0.1}$$\uparrow$}
		& LAB$^\dag$ & 0.532 & 0.235 & 0.495 & 0.543 & 0.539 & 0.449 & 0.463 \\
		& SAN$^\dag$ & 0.536   & 0.236 & 0.462      & 0.555        & 0.552   & 0.456     & 0.493 \\
		& 2 Hourglass$^\dag$ & 0.531   & {\color{blue} \textbf{0.337}} & 0.509      & 0.540        & 0.543   & 0.475     & 0.488 \\
		& DeCaFA$^\dag$ & {\color{blue} \textbf{0.563}} & 0.292 & {\color{blue} \textbf{0.546}}      & {\color{blue} \textbf{0.579}}        & {\color{blue} \textbf{0.575}}   & {\color{blue} \textbf{0.485}}     & {\color{blue} \textbf{0.494}} \\
		& PropNet$^\dag$ & {\color{red} \textbf{0.615}} & {\color{red} \textbf{0.382}} & {\color{red} \textbf{0.628}} & {\color{red} \textbf{0.616}}  & {\color{red} \textbf{0.638}} & {\color{red} \textbf{0.572}}  & {\color{red} \textbf{0.583}} \\ \cline{2-9} 
		& MuSiCa98$^\ddag$ & - & - & - & - & - & - & - \\
		& 3FabRec$^\ddag$ & 0.484   & 0.192 & 0.448      & 0.496        & 0.473   & 0.398     & 0.434 \\
		& $G\&LSR_\omega$$^\ddag$ & 0.493 & - & - & - & - & - & -\\
		& Res18+AVS$^\ddag$ & {\color{blue} \textbf{0.503}}   & {\color{blue} \textbf{0.229}} & {\color{blue} \textbf{0.453}}      & {\color{blue} \textbf{0.525}} & {\color{blue} \textbf{0.484}} & {\color{blue} \textbf{0.431}}     & {\color{blue} \textbf{0.453}} \\
		& ATPN & {\color{red} \textbf{0.557}} & {\color{red} \textbf{0.337}} & {\color{red} \textbf{0.528}} & {\color{red} \textbf{0.568}} & {\color{red} \textbf{0.565}} & {\color{red} \textbf{0.495}} & {\color{red} \textbf{0.516}} \\ \hline
	\end{tabular}
	\caption{Performance comparison of the ATPN and the state-of-the-art methods on WFLW and its subsets (the methods in the table are ranked by the NME on testset). Key: [{\color{red} \textbf{Best}}, {\color{blue} \textbf{Second Best}}, $\downarrow$=the lower the better, $\uparrow$=the larger the better, $\dag$=the method is based on structural information or semi-supervised learning, $\ddag$=the method is for mobile devices]}
	\label{Tabal1}
\end{table*}
Based on the training strategy, head pose estimation and face tracking can take advantages of a part of knowledge of face alignment. We visualize low-level features in the ATPN and two single task CNNs with same layers, as shown in Fig.4. Compared to single task CNNs, the low-level features in the ATPN focus more on the edge of face rather than background. Therefore, sharing low-level features makes CNNs learn more effective features.

\begin{figure}[t!]
	\centering
	\includegraphics[width=\linewidth]{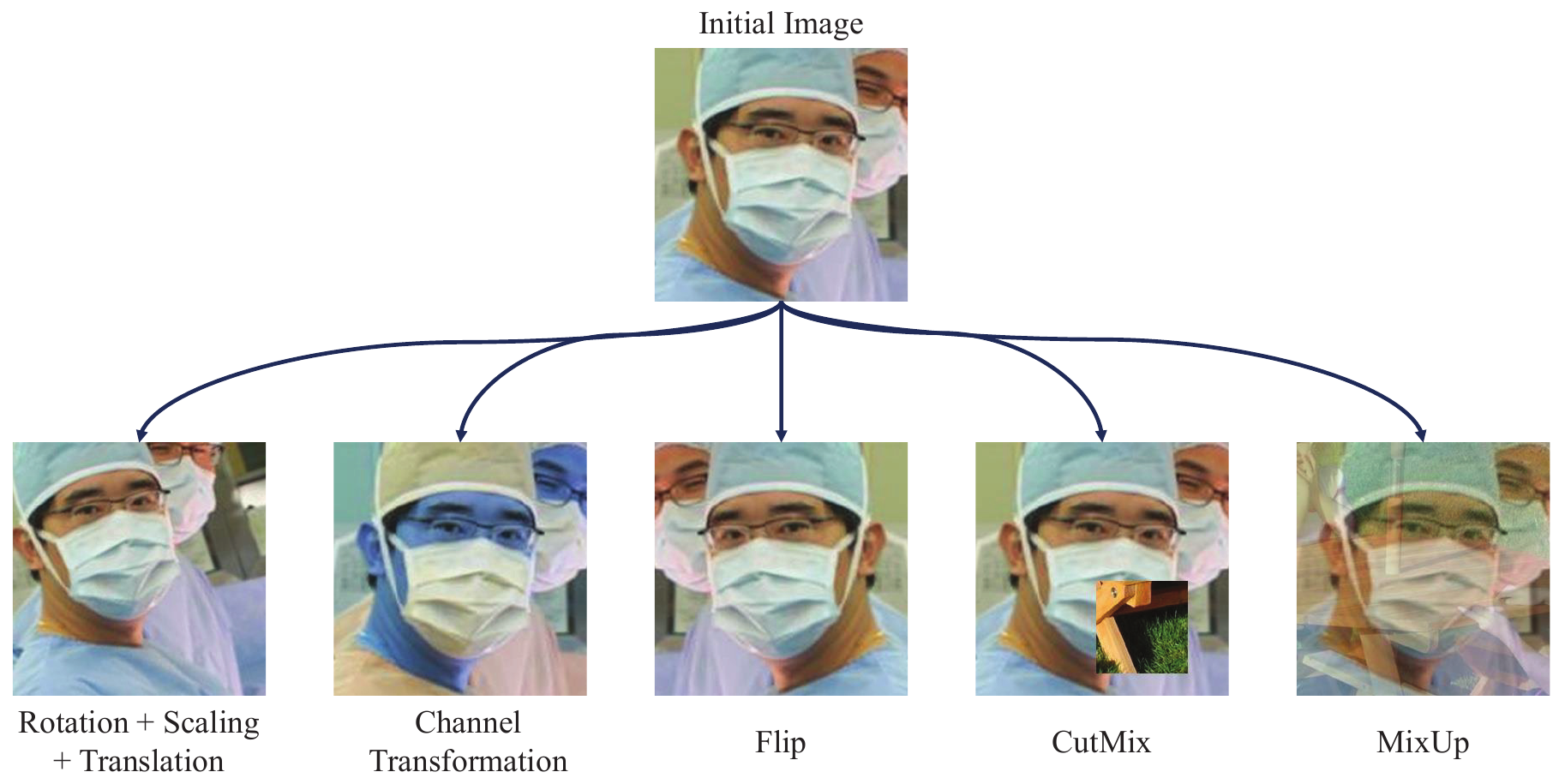}
	\caption{Data augmentations method used in our method.}
	\label{fig5}
\end{figure}

\section{Experiments}

\subsection{Datasets} \label{sec:datasets}

\begin{itemize} 
	\item \textbf{WFLW} \citep{LAB} contains 10,000 faces (7,500 for training and 2,500 for testing) with 98 landmarks. Besides, each face also has attributes annotation in terms of large pose, expression, illumination, make-up, occlusion and blur.
	\item \textbf{300W-LP} \citep{3DDFA} is a synthetic dataset. It fits every sample of 300W \citep{300W} to 3D models by 3DDFA \citep{3DDFA} and rotates 3D model to generate 122450 samples with head pose label.
	\item \textbf{WIDER Face} \citep{WIDER} is a face detection benchmark dataset, including 32,203 images and 393,703 labeled faces with a high degree of variability in scale, pose and occlusion.
	\item \textbf{300VW} \citep{300VW} is a video face alignment dataset, containing 50 videos for training and 64 videos for testing. The testing set is divided into three subsets (category A, B and C).
\end{itemize}

\subsection{Implementation Detail}
We utilize the samples with facial landmark annotation for the first stage training. The learning rate is set to 0.001 and reduced by a factor of 0.03 after every 4 epochs. In the second stage, tracking branch is trained with positive and negative samples created with WIDER Face. We only utilize the face area whose margin is more than 60 for positive samples. For each positive sample generation, 10 negative samples whose IoU is smaller than 0.3 are generated. The learning rate is set to 0.001 and decayed by a factor 0.04 after every 2 epochs. In the third stage, we only use the samples of 300-LP without 3DDFA \citep{3DDFA} data augmentation (3148 for training, 689 for testing) for training. It can keep domain consistency. The method used for data augmentation is the same as that used in the first stage. The learning rate setting is the same as the second stages.

Moreover, each sample randomly translates, rotates, flips, transforms channels or mixes up with the background image created by WIDER to augment the training data, as shown in Fig.5. we utilize Adam \citep{Adam} optimizer with a $\beta_1$ of 0.9 and a $\beta_2$ of 0.999 in the three stages. The batch size is set to 128.

\subsection{Evaluation Metrics}
\subsubsection{Face Alignment}
We evaluate the proposed method with Normalized Mean Error (NME). In WFLW, we use the inter-ocular distance \citep{300W} as the normalization factor. In 300VW, we also use the inter-pupil distance \citep{LAB} as the normalization factor to compare ATPN with other methods. Besides, we also report the Failure Rate (FR) for a maximum error of 0.1 and Area Under Curve (AUC) derived from Cumulative Errors Distribution (CED) curve by plotting the curve from 0 to 0.1.

\subsubsection{Head Pose Estimation}
Following the works of \citep{LAB,PropNet}, we divide the test set into two subsets (common subset and challenging subset) and report the Mean Average Error (MAE), as given below:
\begin{equation}
MAE = \frac{1}{{{N_{test}}}}\sum\limits_{l = 1}^{N{}_{test}} {\frac{{\left| {{\alpha _l} - {\alpha _l}^\prime } \right| + \left| {{\beta _l} - {\beta _l}^\prime } \right| + \left| {{\gamma _l} - {\gamma _l}^\prime } \right|}}{3}},
\end{equation}
where $(\alpha,\beta,\gamma)$ and $(\alpha^\prime,\beta^\prime,\gamma^\prime)$ are respectively the predicted and annotated roll, pitch, yaw angle of head. $N_{test}$ is the total number of testing samples. Besides, we also report the FR in for a maximum error of $10^\circ$ and CED curve.
\subsubsection{Face Tracking}
We evaluate ATPN with the NME and tracking failure rate. The threshold of failure rate is set to 0.7. Besides, we also report the Precision-Recall (PR) curve and Average Precision (AP) on the test set of WIDER Face.

\subsection{Comparison with State-of-The-Art Methods}
\subsubsection{WFLW}
Table 1 compares the ATPN to the methods based on structural information and the methods for mobile devices respectively. Table 2 illustrates number of parameters and computational complexity of the methods. Despite the PropNet achieves best performance of 4.05\% in NME, the computational complexity is much more than the methods for mobile devices (more than 150$\times$). By taking advantages of the structural information at low-level features, ATPN achieves best performance in NME among the methods for mobile devices. Compared to $G\&LSR_\omega$, we observe an improvement of 2.47\% in NME with comparable computational complexity. Although LAB adopts GAN to generate a boundary heatmap, ATPN still outperforms it with only 0.3\% computational complexity and 8.9\% parameters. Therefore, ATPN is much more efficient than the state-of-the-art methods.
%is Compared to other methods, LAB, DeCaFA and PropNet are based on Hourglass module and can utilize structural information to localize facial landmarks. LAB utilizes two hourglass modules to generate a facial boundary heatmap and fuses it with the input image, while DeCaFA and PropNet utilize four modules. Table 2 shows the number of parameters of these methods. ATPN obtains the least computational complexity among all methods, only 0.3\% of LAB. But it achieves a 2.6\% performance gain compared to LAB and a 1.2\% gain to 2 Hourglass. The main reason is that regressing facial landmarks from image directly is much easier than generating a heatmap. Besides, the boundary heatmap may lead to additional error in CNNs. The ATPN can also acquire structural information of face by low-level features, which guarantees the accuracy. However, Hourglass module enables model to utilize more cascaded modules to achieve better result. As a result, those models based on four Hourglass modules, such as PropNet and DeCaFA, demonstrate better performance. But it makes models less efficient.

%The main reason is that it does not require hourglass module to acquire the structural information. Moreover, the parameters of the ATPN can be reduced further if it does not estimate head pose. Besides, all methods mentioned in Table 1 require a face detection procedure before processing and that is why these methods cannot run as fast as the speed in theory. 

\begin{table}[t!]
	\centering
	\begin{tabular}{m{3cm}<{\centering}m{1.6cm}<{\centering}m{1.6cm}<{\centering}}
		\hline
		Method  & Params(M)$\downarrow$ & FLOPS(G)$\downarrow$ \\ \hline
		PropNet \citep{PropNet} & 36.3  & 42.83 \\ 
		LAB\citep{LAB} & 12.3 & 18.85 \\
		DeCaFA \citep{DeCaFA} & $\approx$10 & $\approx$30  \\
		2 Hourglass \citep{Hourglass} & 6.30 & 8.00 \\
		MuSiCa98$^\ddag$ \citep{KWTQ} & $\approx$3 & {\color{blue} \textbf{$\approx$0.25}}\\
		$G\&LSR_\omega$ \citep{RFA} & {\color{blue} \textbf{1.83}} & {\color{red} \textbf{0.06}} \\
		ATPN (Backbone + Alignment Branch) & {\color{red} \textbf{1.1}} & {\color{red} \textbf{0.06}} \\ \hline
	\end{tabular}
	\caption{Parameters and computational complexity for ATPN and other the state-of-the-art methods (the methods in the table are ranked by the number of parameters). Key: [{\color{red} \textbf{Best}}, {\color{blue} \textbf{Second Best}}, $\downarrow$=the lower the better]}
	\label{Tabal2}
\end{table}

\begin{table}[t!]
	\centering
	\begin{tabular}{cccc}
		\hline
		\multicolumn{1}{m{3.5cm}<{\centering}|}{Method} & \multicolumn{1}{c|}{Cat A} & \multicolumn{1}{c|}{Cat B} & Cat C \\ \hline
		\multicolumn{4}{c}{inter-ocular distance normalization (\%)$\downarrow$}  \\ \hline
		\multicolumn{1}{m{3.5cm}<{\centering}|}{TSCN \citep{Two-Stream-action}}             & \multicolumn{1}{c|}{12.54}      & \multicolumn{1}{c|}{7.25}       & 13.13      \\
		\multicolumn{1}{m{3.5cm}<{\centering}|}{CFSS \citep{CFSS}} & \multicolumn{1}{c|}{7.68} & \multicolumn{1}{c|}{6.42} & 13.67  \\
		\multicolumn{1}{m{3.5cm}<{\centering}|}{SDM \citep{SDM}} & \multicolumn{1}{c|}{7.41} & \multicolumn{1}{c|}{6.08}  & 14.03 \\
		\multicolumn{1}{m{3.5cm}<{\centering}|}{TSTN \citep{Two-stream_video}}             & \multicolumn{1}{c|}{5.21}       & \multicolumn{1}{c|}{4.23}       & 10.11      \\
		\multicolumn{1}{m{3.5cm}<{\centering}|}{ADC \citep{ADC}}              & \multicolumn{1}{c|}{4.17}       & \multicolumn{1}{c|}{3.89}       & 7.28       \\ 
		\multicolumn{1}{m{3.5cm}<{\centering}|}{DeCaFA \citep{DeCaFA}}           & \multicolumn{1}{c|}{3.82}       & \multicolumn{1}{c|}{3.63}       & 6.67       \\
		\multicolumn{1}{m{3.5cm}<{\centering}|}{FAB \citep{FAB}}              & \multicolumn{1}{c|}{3.56}       & \multicolumn{1}{c|}{3.88}       & 5.02       \\
		\hline
		\multicolumn{1}{m{3.5cm}<{\centering}|}{ATPN (Tracking)}  & \multicolumn{1}{c|}{{\color{blue} \textbf{3.52}}}       & \multicolumn{1}{c|}{{\color{blue} \textbf{3.64}}}       & {\color{blue} \textbf{4.99}}       \\
		\multicolumn{1}{m{3.5cm}<{\centering}|}{ATPN (Detection)} & \multicolumn{1}{c|}{{\color{red} \textbf{3.49}}} & \multicolumn{1}{c|}{{\color{red} \textbf{3.57}}}       & {\color{red} \textbf{4.89}}       \\ \hline
		\multicolumn{4}{c}{inter-pupil distance normalization (\%)$\downarrow$}                                                                             \\ \hline
		\multicolumn{1}{m{3.5cm}<{\centering}|}{4S-HG \citep{Hourglass}}            & \multicolumn{1}{c|}{6.54}       & \multicolumn{1}{c|}{5.65}       & 8.13       \\
		\multicolumn{1}{m{3.5cm}<{\centering}|}{4S-HG+UFLD \citep{TUF}}       & \multicolumn{1}{c|}{6.09}       & \multicolumn{1}{c|}{5.34}       & 7.76       \\ \hline
		\multicolumn{1}{m{3.5cm}<{\centering}|}{ATPN (Tracking)}  & \multicolumn{1}{c|}{{\color{blue} \textbf{4.83}}}       & \multicolumn{1}{c|}{{\color{blue} \textbf{5.04}}}       & {\color{blue} \textbf{7.75}}       \\
		\multicolumn{1}{m{3.5cm}<{\centering}|}{ATPN (Detection)} & \multicolumn{1}{c|}{{\color{red} \textbf{4.79}}}       & \multicolumn{1}{c|}{{\color{red} \textbf{4.95}}}       & {\color{red} \textbf{7.61}}       \\ \hline
	\end{tabular}
	\caption{NME for the ATPN in tracking and detection mode compared with previous methods on Category A, Category B and Category C of 300VW (the methods in the table are ranked by the NME on Category A). Key: [{\color{red} \textbf{Best}}, {\color{blue} \textbf{Second Best}}, $\downarrow$=the lower the better]}
	\label{Tabal3}
\end{table}

\subsubsection{300VW}
We pretrain the ATPN on WFLW and then retrain it with the training set of 300VW. We carry out two different experiments on three subsets of 300VW. In the frist one, we evaluate ATPN with detection mode. In this mode, the input image is cropped based on the ground truth. In the second experiment, the input image is cropped based on the face alignment result in the last frame for tracking. The comparison result are illustrated in Table 3. The most samples in 300VW are without occlusion, which enables the low-level features to produce more complete facial boundaries. Therefore, the improvement of ATPN is more significantly compared to other state-of-the-art methods. For example, although DeCaFA achieves excellent performance on WFLW, ATPN still achieves an impressive improvement of 8.64\%, 1.66\% and 26.69\% in NME respectively on the three subsets

Moreover, the tracking failure rates are only 0.14\%, 0.00\% and 0.08\% on the three subsets, which means the tracking branch is with satisfactory robustness. As a result, the performance of tracking mode only degrades a little compared to detection mode. Besides, in Fig.6, we also demonstrate the PR curves of the ATPN and Hyperface \citep{Hyperface} on the validation set of WIDER Face. The average precision of ATPN reaches at 98.82\% that outperforms Hyperface a lot. Some predicted results in the validation set can be viewed in Fig.7.

The Table 5 shows the parameters and computational complexity of the tracking branch and a commonly used face detection framework (MTCNN). The MTCNN is with $10\times$ higher computational complexity than ATPN. Hence, the real-time capability of ATPN will degrades dramatically if the face detection framework is employed in each frame. By taking advantages of tracking branch, the processing time can be accelerated more than 50 times.
%In the detection mode, the ROI of face is extracted based on the ground truth in each frame. In the tracking mode, we generate ROI using the face alignment result of the last frame unless it is the first frame of a sequence or face confidence is less than 0.7. The NME of ATPN and state-of-the-art methods are shown in Table 3 and tracking failure rate is shown in Table 4. Compared to detection mode, NME of tracking mode only increases slightly while it does not require face detection in most of frames. The tracking failure rate is only 0.14\% in Category A and 0.08 \% in Category C. The experimental result illustrates the tracking branch is quite reliable. And the number of parameters and computational complexity of the tracking branch and the face detector (MTCNN \citep{MTCNN}) are shown in Table 5. The computational complexity of the tracking branch is only 1.2\%$ \sim $2.4\% of the face detector. Therefore, the tracking branch can improve the efficiency of the model significantly. 

\begin{table}[t!]
	\centering
	\begin{tabular}{|l|l|l|l|}
		\hline
		& Cat A & Cat B & Cat C \\ \hline
		Sequences       & 146        & 39         & 188        \\ \hline
		Detection frame & 233        & 39         & 210        \\ \hline
		Tracking frame  & 61902      & 32766      & 26128      \\ \hline
		Failure frame   & 87         & 0          & 22         \\ \hline
		Failure rate    & 0.14\%     & 0.00\%     & 0.08\%     \\ \hline
	\end{tabular}
	\caption{Tracking details on 300VW. \textbf{Note}: 300VW fails to label some frames in several videos, resulting in a video being divided into dozens of sequences.}
	\label{Tabal4}
\end{table}

\begin{table}[t!]
	\centering
	\begin{tabular}{m{3.0cm}<{\centering}m{1.6cm}<{\centering}m{1.6cm}<{\centering}}
		\hline
		Method          & Params(M)$\downarrow$ & FLOPS(G)$\downarrow$ \\ \hline
		MTCNN \citep{MTCNN}   & 0.47  & 0.7$ \sim $1.4  \\ 
		Tracking Branch & {\color{red} \textbf{0.19}} & {\color{red} \textbf{0.017}}    \\ \hline
	\end{tabular}
	\caption{Parameters and computational complexity for tracking branch and MTCNN (the methods in the table are ranked by the number of parameters). Key: [{\color{red} \textbf{Best}}, $\downarrow$=the lower the better].}
	\label{Tabal5}
\end{table}

\begin{table*}[t!]
	\centering
	\begin{tabular}{m{4.2cm}<{\centering}|m{0.6cm}<{\centering}m{0.7cm}<{\centering}m{0.6cm}<{\centering}m{0.7cm}<{\centering}m{1.0cm}<{\centering}|m{0.6cm}<{\centering}m{0.7cm}<{\centering}m{0.6cm}<{\centering}m{0.7cm}<{\centering}m{1.0cm}<{\centering}}
		\hline
		\multirow{2}{*}{Method} & \multicolumn{5}{c|}{Common Subset}   & \multicolumn{5}{c}{Challenging Subset}  \\ \cline{2-11} 
		& Yaw$\downarrow$  & Pitch$\downarrow$ & Roll$\downarrow$ & MAE$\downarrow$  & FR$_{10^\circ}$$\downarrow$   & Yaw$\downarrow$   & Pitch$\downarrow$ & Roll$\downarrow$  & MAE$\downarrow$   & FR$_{10^\circ}$$\downarrow$   \\ \hline
		ESR \citep{ESR}$^\star$ & 8.54$^\circ$ & 7.47$^\circ$  & 2.61$^\circ$ & 6.21$^\circ$ & 19.31\% & 24.52$^\circ$ & 12.70$^\circ$ & 10.64$^\circ$ & 15.95$^\circ$ & 50.62\% \\
		Annotated landmarks$^\star$   & 7.32$^\circ$ & 6.99$^\circ$  & 1.88$^\circ$ & 5.40$^\circ$ & 16.37\% & 11.64$^\circ$ & 9.96$^\circ$  & 5.40$^\circ$  & 9.00$^\circ$  & 32.84\% \\  
		Hyperface \citep{Hyperface} & 3.46$^\circ$ & 3.87$^\circ$  & 2.92$^\circ$ & 3.42$^\circ$ & 3.37\%  & 6.30$^\circ$  & 5.57$^\circ$  & 7.34$^\circ$  & 6.40$^\circ$  & 19.76\% \\
		Openface2.0 \citep{Openface} & 2.69$^\circ$ & 3.56$^\circ$  & {\color{blue} $\bm{1.10^\circ}$} & 2.45$^\circ$ & {\color{blue} $\bm{1.72\%}$}  & {\color{blue} $\bm{3.23^\circ}$}  &{\color{red}  $\bm{3.56^\circ}$}  & {\color{red} $\bm{1.93^\circ}$}  & {\color{blue} $\bm{2.91^\circ}$}  & {\color{blue} $\bm{1.51\%}$}  \\
		FSA-Net \citep{FSA-Net} & {\color{blue} $\bm{1.85^\circ}$} & {\color{blue} $\bm{2.84^\circ}$}  & 1.12$^\circ$ & {\color{blue} $\bm{1.94^\circ}$} & {\color{blue} $\bm{2.35\%}$}  & 4.51$^\circ$  & 5.29$^\circ$  & {\color{blue} $\bm{2.27^\circ}$}  & 4.12$^\circ$  & 19.26\% \\
		ATPN  & {\color{red} $\bm{1.31^\circ}$} & {\color{red} $\bm{2.38^\circ}$}  & {\color{red} $\bm{0.97^\circ}$} & {\color{red} $\bm{1.55^\circ}$} & {\color{red} $\bm{0.12\%}$}  & {\color{red} $\bm{2.62^\circ}$}  & {\color{blue} $\bm{3.97^\circ}$}  & {\color{red} \bm{$1.93^\circ$}}  & {\color{red} $\bm{2.84^\circ}$}  & {\color{red} $\bm{1.49\%}$}  \\ \hline
	\end{tabular}
	\caption{Mean Average Error (MAE) in degrees and FR$_{10^\circ}$ on the common and challenging subset (the methods in the table are ranked by the MAE on the common subset). Key: [{\color{red} \textbf{Best}}, {\color{blue} \textbf{Second Best}}, $\downarrow$=the lower the better, $^\star$=5-landmarks model-based method (eyes corner, mouth corner, nose tip and chin)]}
	\label{Tabal6}
\end{table*}

\begin{table*}[t!]
	\centering
	\begin{tabular}{m{1.05cm}<{\centering}|m{1.4cm}<{\centering}m{1.1cm}<{\centering}m{1.4cm}<{\centering}|m{0.6cm}<{\centering}m{0.6cm}<{\centering}m{0.6cm}<{\centering}m{0.6cm}<{\centering}m{0.6cm}<{\centering}m{0.6cm}<{\centering}m{0.6cm}<{\centering}|c}
		\hline
		\multirow{3}{*}{Method} & \multicolumn{3}{c|}{Modules} & \multicolumn{7}{c|}{NME(\%)$\downarrow$} & FR$_{0.1}$(\%)$\downarrow$ \\ \cline{2-12} 
		& low-level Features & Multiview Block & CoordConv & Testset & Pose & Expre-ssion & Illumi-nation & Make-up & Occlu-sion & Blur & Testset \\ \hline
		Baseline & & &  & 5.61    & 9.90 & 6.11       & 5.36         & 5.47    & 6.77      & 6.38 & 8.68       \\
		Model1                  &    \checkmark          &                 &           & 5.40    & 9.22 & 5.92       & 5.30         & 5.30    & 6.43      & 6.00 & 7.16       \\
		Model2                  &     \checkmark        &   \checkmark      &           & 5.33    & 9.15 & 5.75       & 5.16         & 5.33    & 6.36      & 5.95 & 7.52       \\
		Model3                  &    \checkmark      &   \checkmark  & \checkmark  & {\color{red} \textbf{$\bm{5.13}$}}    & {\color{red} \textbf{$\bm{8.97}$}} & {\color{red} \textbf{$\bm{5.49}$}} & {\color{red} \textbf{$\bm{4.95}$}} & {\color{red} \textbf{$\bm{4.94}$}}    & {\color{red} \textbf{$\bm{6.30}$}} & {\color{red} \textbf{$\bm{5.78}$}} & {\color{red} \textbf{$\bm{6.27}$}}    \\ \hline
	\end{tabular}
	\caption{The contribution of different modules (the methods in the table are ranked by the NME in the test set). Key:[{\color{red} \textbf{Best}}, $\downarrow$=the lower the better]}
	\label{Tabal7}
\end{table*}

\begin{figure}[t!]
	\centering
	\includegraphics[width=\linewidth]{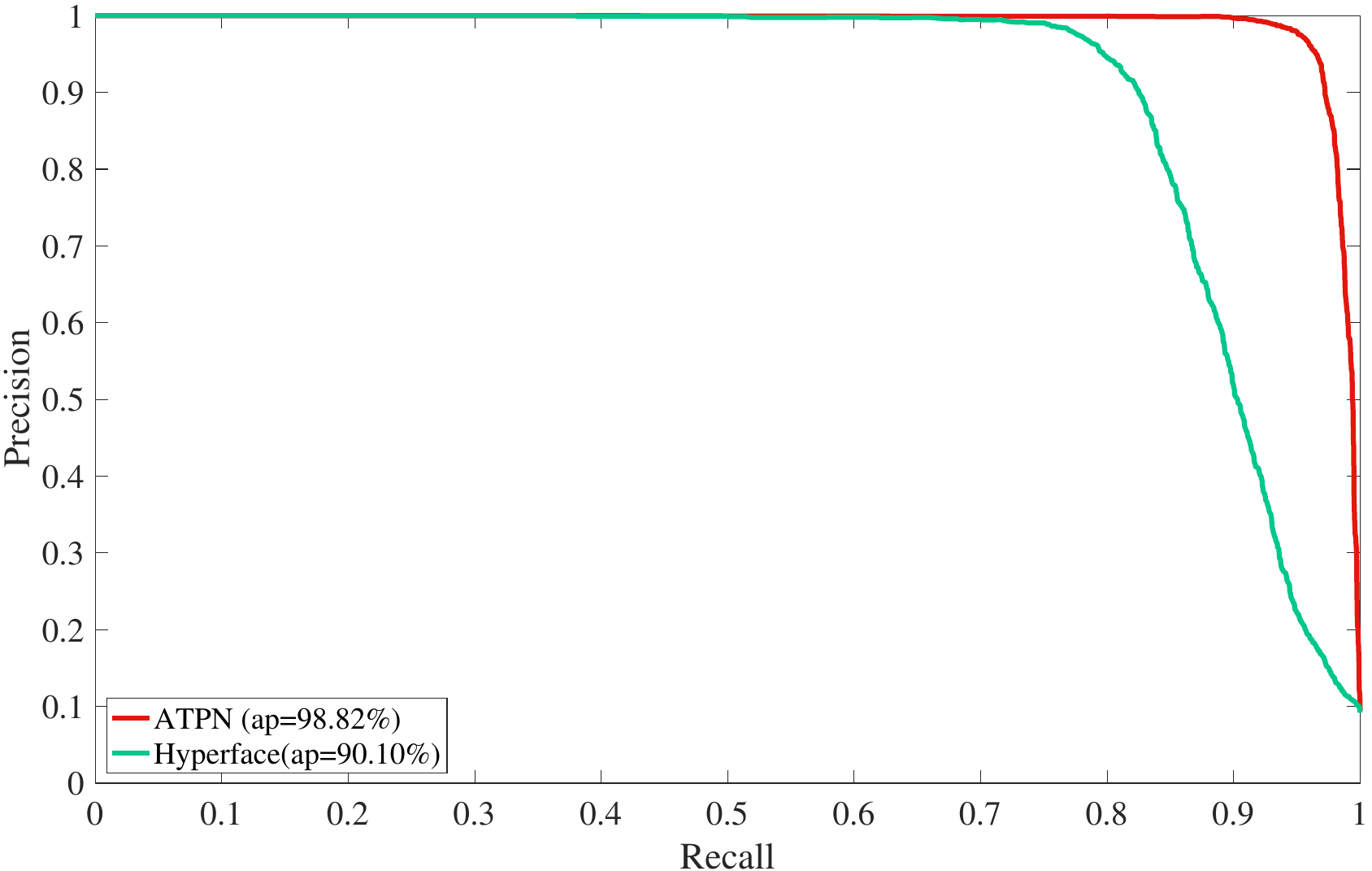}
	\caption{Evaluation on the validation set of WIDER FACE. The number following the method indicates the average accuracy of face tracking.}
	\label{fig6}
\end{figure}

\subsubsection{300W-LP}
We compare the pose branch to other state-of-the-arts methods on both the common and challenging subset, as shown in Table 6. Besides, Fig.8 also illustrates the CED curves of the ATPN and other methods on the full set. It is difficult for 5-landmarks model-based method to achieve accurate head pose estimation even if it utilizes the annotated landmarks. Hyperface and FSA-Net utilize the appearance features of the input image. The performance is improved significantly because the appearance features contains more information compared to 5 landmarks. Nevertheless, it degrades significantly for the images with extreme conditions (challenging subset) because of the fragile robustness. OpenPose2.0 adopt 3D geometric information to estimate head pose and it can maintain excellent performance on the challenging subset. ATPN estimates head pose based on both geometric and appearance features, which significantly reduces the MAE by 36.73\%, from $2.45^\circ$ to $1.55^\circ$ on the common subset compared to Openface2.0. Besides, the geometric features enables ATPN to maintain comparable performance.

\begin{figure*}[t!]
	\centering
	\includegraphics[width=\linewidth]{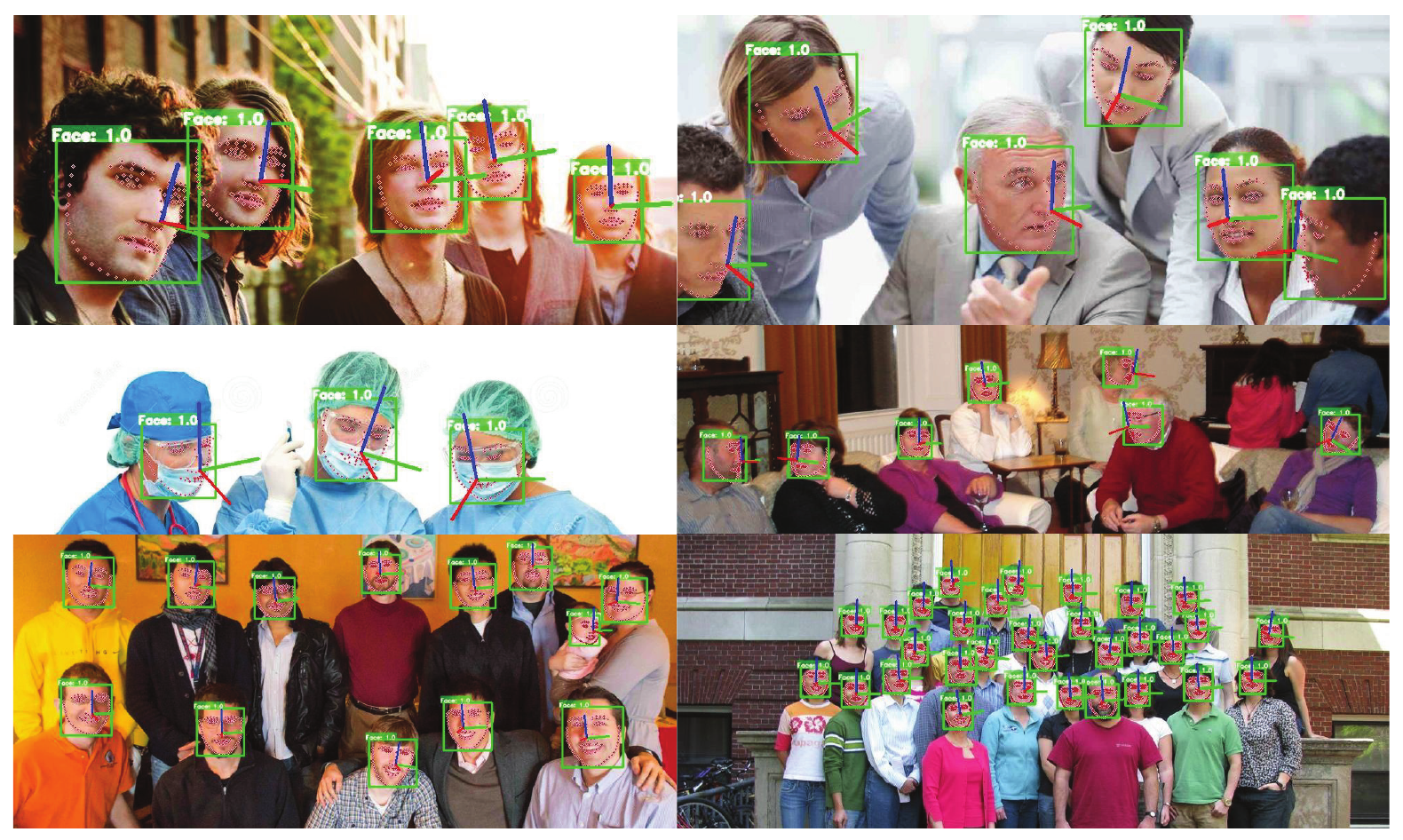}
	\caption{Sample results on WIDER Face validation set (the number above the bounding box indicates the confidence of face, the {\color{red} \textbf{red axis}} points towards the front of the face, {\color{blue} \textbf{blue}} pointing upward and {\color{green} \textbf{green}} pointing left side).}
	\label{fig7}
\end{figure*}

\subsection{Ablation Study}

\begin{table*}[t!]
	\centering
	\begin{tabular}{c|m{0.6cm}<{\centering}m{0.7cm}<{\centering}m{0.6cm}<{\centering}m{0.7cm}<{\centering}m{1.2cm}<{\centering}|m{0.6cm}<{\centering}m{0.7cm}<{\centering}m{0.6cm}<{\centering}m{0.7cm}<{\centering}m{1.2cm}<{\centering}}
		\hline
		\multirow{2}{*}{Method} & \multicolumn{5}{c|}{Common Subset}   & \multicolumn{5}{c}{Challenging Subset}  \\ \cline{2-11} 
		& Yaw$\downarrow$  & Pitch$\downarrow$ & Roll$\downarrow$ & MAE$\downarrow$  & FR$_{10^\circ}$$\downarrow$   & Yaw$\downarrow$   & Pitch$\downarrow$ & Roll$\downarrow$  & MAE$\downarrow$   & FR$_{10^\circ}$$\downarrow$   \\ \hline
		Baseline                    & $1.47^\circ$ & $2.53^\circ$  & $1.03^\circ$ & $1.68^\circ$ & $0.12\%$  & $2.93^\circ$  & 3.98$^\circ$  & $2.44^\circ$  & $3.12^\circ$  & $4.70\%$  \\
		Feature sharing & 1.42$^\circ$ & 2.60$^\circ$  & 1.11$^\circ$ & 1.71$^\circ$ & 0.24\%  & 2.96$^\circ$  & $4.17^\circ$  & 2.38$^\circ$  & 3.17$^\circ$  & 3.46\%  \\
		Feature sharing+Heatmap    & {\color{red} \textbf{$\bm{1.31^\circ}$}} & {\color{red} \textbf{$\bm{2.38^\circ}$}}  & {\color{red} \textbf{$\bm{0.97^\circ}$}} & {\color{red} \textbf{$\bm{1.55^\circ}$}} & {\color{red} \textbf{$\bm{0.12\%}$}} & {\color{red} \textbf{$\bm{2.62^\circ}$}} & {\color{red} \textbf{$\bm{3.97^\circ}$}} & {\color{red} \textbf{$\bm{1.93^\circ}$}} & {\color{red} \textbf{$\bm{2.84^\circ}$}} & {\color{red} \textbf{$\bm{1.49\%}$}} \\ \hline
	\end{tabular}
	\caption{The effects of feature sharing and heatmap on face tracking on common subset and challenging subset (the methods in the table are ranked by the MAE). Key: [{\color{red} \textbf{Best}}, $\downarrow$=the lower the better]}
	\label{Tabal9}
\end{table*}

\begin{table}[t!]
	\centering
	\begin{tabular}{m{1.6cm}<{\centering}|m{1.1cm}<{\centering}m{1.1cm}<{\centering}m{2.3cm}<{\centering}}
		\hline
		Method  & Baseline & Feature Sharing & Feature Sharing + Heatmap \\ \hline
		Average Precision $\uparrow$ & 98.57\%  & 98.72\% & {\color{red} \textbf{$\bm{98.82\%}$}} \\ \hline
	\end{tabular}
	\caption{The influence of feature sharing and heatmap on the face tracking on WIDER Face. Key: [{\color{red} \textbf{Best}}, $\uparrow$=the larger the better]}
	\label{Tabal8}
\end{table}

\subsubsection{Alignment Branch}
The alignment branch consists of several pivotal modules: the structural information at low-level, multiview block and CoordConv. We compare a baseline with several models that consist of these modules. The baseline is set as a model that regresses facial landmarks directly from the last layer of the backbone. The evaluation results on WFLW are shown in Table 7.

%To study the effectiveness of the modules, we set a baseline as a model which regresses facial landmarks directly from the last layer of the backbone network, and then we compare it with other models consist of these modules. The evaluation results on WFLW are shown in Table 7.

\begin{figure}[t!]
	\centering
	\includegraphics[width=\linewidth]{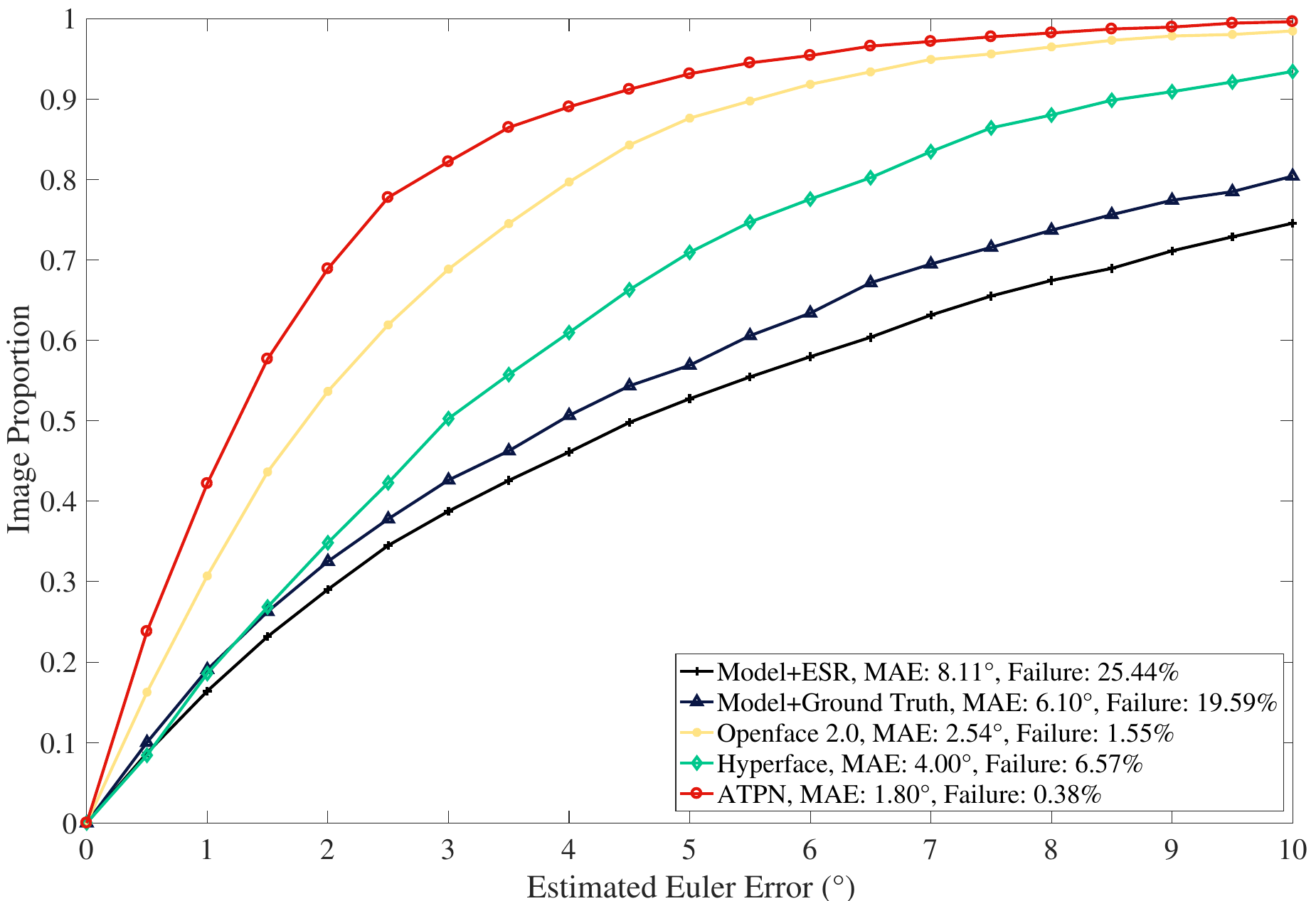}
	\caption{CED for the full set (Yaw, Pitch, Roll). MAE and Failure Rate are also reported.}
	\label{fig8}
\end{figure}

\begin{figure}[t!]
	\centering
	\includegraphics[width=\linewidth]{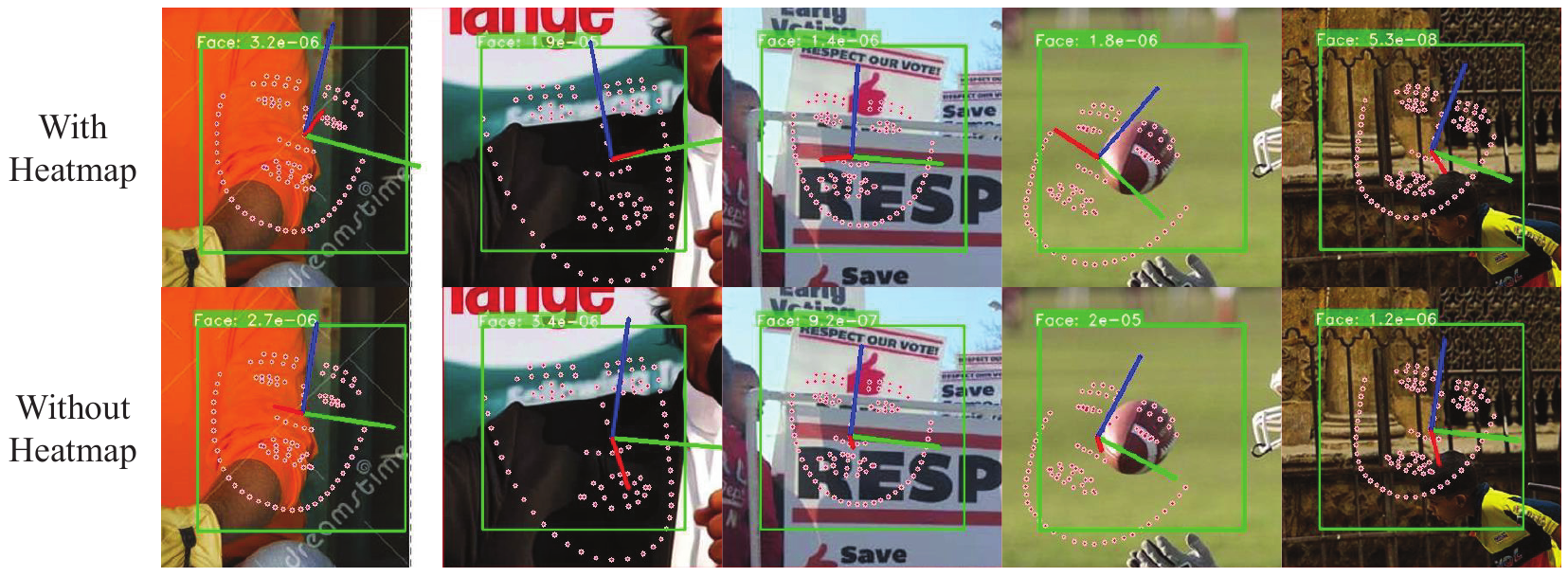}
	\caption{Estimation results of the model with/without heatmap on the background images (the {\color{red} \textbf{red axis}} points towards the front of the face, {\color{blue} \textbf{blue}} pointing upward and {\color{green} \textbf{green}} pointing left side).}
	\label{fig9}
\end{figure}

We observe 3.74\% and 17.51\% improvement respectively in NME and FR$_{0.1}$ by introducing the structural information into CNN. It illustrates that the structural information of low-level features is essential to face alignment. Then, CoordConv also improves the performance of face alignment significantly by providing spatial information for ATPN. Without adding much computational complexity, NME and FR are reduced by 3.57\% and 16.62\% respectively. Finally, the improvement brought by multiview block is not as significant as others, although it obtains the largest number of parameters. It suggests that increasing parameters in network cannot improve the performance effectively. 

\subsubsection{Feature Sharing and Heatmap}
Feature sharing and heatmap also play an important role in the tracking and pose branch. To further investigate whether the heatmap can introduce the geometric information into the network, we carry out an additional experiment on background images and some results are shown in Fig.9. Despite the input images do not include any face, the predicted head pose of the model with heatmap is still highly correspond to the predicted facial landmarks, while the results of the model without heatmap are not. It illustrates that when the appearance features are not reliable, the model with heatmap can utilize the geometric information for head pose estimation.

To study the influence of feature sharing and heatmap on two tasks, we set the baseline as a single task CNN with the same layers. In feature sharing model, the low-level layers are shared with face alignment and only high-level layers are trained.

The results of head pose estimation on 300W-LP are shown in Table 8. Compared to the baseline, the feature sharing model exhibits the comparable performance on the both common and challenging subset with much less parameters. With the geometric information of the heatmap, the MAE is significantly improved by 10.41\% and 9.36\% respectively on the common and challenging subset. Therefore, sharing features can boost the real-time capability and the geometric information of the heatmap is crucial to head pose estimation.

In terms of face tracking, the experiment is carried out on WIDER Face. The influence of feature sharing and heatmap is shown in Table 9. Although the trainable parameters in features sharing model is much less than the baseline, the average precision is still improved from 98.57\% to 98.72\%. By introducing the attention clue into network with heatmap, the performance is further boosted to 98.82\%. It indicates that the features learned by face alignment task and the attention clue of heatmap can improve the performance of face tracking.

\section{CONCLUSION}
In this paper, we present the Alignment \& Tracking \& Pose Network (ATPN), an efficient multitask network for facial landmark localization, face tracking and head pose estimation. Different from other state-of-the-art works that generates the structural information by an additional network or branch, we directly utilize the structural information in the low-level features. By taking advantages of the structural information, the computational complexity is only $1/300$ of the other methods that utilize the structural information. Moreover, it also achieves the best performance with the least parameters and FLOPS compared to other methods for mobile devices. On the video dataset, the improvement of real-time capability is more significant because the tracking branch eliminates the face detection process. Different from other multitask frameworks that only share features, ATPN also generates a heatmap by face alignment result to provide the geometric information for head pose estimation and the attention clues for face tracking. The experiment results illustrate that the geometric information of heatmap can help network maintain robustness on the extreme conditions. And the attention clues provided by heatmap can also improve the accuracy of face tracking.

In our future work, we will further explore the features sharing mechanism and attention mechanism learned by semi-supervised learning. Based on the mechanisms, we will develop a lighter model to boost both the performance and real-time capability of the state-of-the-art methods and apply it on the mobile devices.  
%This paper presents ATPN, a practical multitask CNN for face alignment, tracking and head pose estimation. First, different from previous studies, this framework utilizes structural information of face in shallow layers to achieve high accuracy with less parameters. Second, a cheap heatmap is generated directly from face alignment result and then it is fused with features to make CNN utilize both geometric and appearance information to estimate head pose. The robustness of face tracking is also improved by the heatmap by acquiring attention clues. Third, the tracking branch enables ATPN to remain excellent performance in video and achieve better real-time performance. The experimental results show the ATPN obtains better comprehensive ability compared to privious state-of-the-art methods and our designs are effective to enhance three tasks' performance. In our future work, we would investigate combining our model with semi-supervised learning to boost performance further.

\newpage

\section*{Appendix A}
All abbreviations used in this paper are described in Table 10.

\begin{table}[h]
	\centering
	\begin{tabular}{m{1.5cm}<{\centering}m{5.6cm}<{\centering}}
		\hline
		Abbreviation & Description                           \\ \hline
		CNN          & Convolutional Neural Network          \\
		ATPN         & Alignment \& Tracking \& Pose Network \\
		ROI          & Region of Interest                    \\
		NME          & Normalized Mean Error                 \\
		FR           & Failure Rate                          \\
		AUC          & Area Under Curve                      \\
		CED          & Cumulative Errors Distribution        \\
		MAE          & Mean Average Error                    \\
		PR           & Precision-Recall                      \\
		AP           & Average Precision                     \\ \hline
	\end{tabular}
	\caption{Lookup table for abbreviations in the paper.}
\end{table}

\section*{Acknowledgement}
We would like to thank the anonymous reviewers for reviewing this manuscript.

\bibliography{ref.bib}

\begin{thebibliography}{52}
\expandafter\ifx\csname natexlab\endcsname\relax\def\natexlab#1{#1}\fi
\providecommand{\url}[1]{\texttt{#1}}
\providecommand{\href}[2]{#2}
\providecommand{\path}[1]{#1}
\providecommand{\DOIprefix}{doi:}
\providecommand{\ArXivprefix}{arXiv:}
\providecommand{\URLprefix}{URL: }
\providecommand{\Pubmedprefix}{pmid:}
\providecommand{\doi}[1]{\href{http://dx.doi.org/#1}{\path{#1}}}
\providecommand{\Pubmed}[1]{\href{pmid:#1}{\path{#1}}}
\providecommand{\bibinfo}[2]{#2}
\ifx\xfnm\relax \def\xfnm[#1]{\unskip,\space#1}\fi
%Type = Article
\bibitem[{Ahn et~al.(2018)Ahn, Choi, Park \& Kweon}]{Multitask-pose}
\bibinfo{author}{Ahn, B.}, \bibinfo{author}{Choi, D.-G.},
  \bibinfo{author}{Park, J.}, \& \bibinfo{author}{Kweon, I.~S.}
  (\bibinfo{year}{2018}).
\newblock \bibinfo{title}{Real-time head pose estimation using multi-task deep
  neural network}.
\newblock {\it \bibinfo{journal}{Robotics and Autonomous Systems}\/},  {\it
  \bibinfo{volume}{103}\/}, \bibinfo{pages}{1--12}.
  \DOIprefix\doi{https://doi.org/10.1016/j.robot.2018.01.005}.
%Type = Article
\bibitem[{Bakker et~al.(2021)Bakker, Zabłocki, Baker, Riethmeister, Marx,
  Iyer, Anund \& Ahlström}]{fatigue}
\bibinfo{author}{Bakker, B.}, \bibinfo{author}{Zabłocki, B.},
  \bibinfo{author}{Baker, A.}, \bibinfo{author}{Riethmeister, V.},
  \bibinfo{author}{Marx, B.}, \bibinfo{author}{Iyer, G.},
  \bibinfo{author}{Anund, A.}, \& \bibinfo{author}{Ahlström, C.}
  (\bibinfo{year}{2021}).
\newblock \bibinfo{title}{A multi-stage, multi-feature machine learning
  approach to detect driver sleepiness in naturalistic road driving
  conditions}.
\newblock {\it \bibinfo{journal}{IEEE Transactions on Intelligent
  Transportation Systems}\/},  (pp. \bibinfo{pages}{1--10}).
  \DOIprefix\doi{10.1109/TITS.2021.3090272}.
%Type = Inproceedings
\bibitem[{{Baltrusaitis} et~al.(2018){Baltrusaitis}, {Zadeh}, {Lim} \&
  {Morency}}]{Openface}
\bibinfo{author}{{Baltrusaitis}, T.}, \bibinfo{author}{{Zadeh}, A.},
  \bibinfo{author}{{Lim}, Y.~C.}, \& \bibinfo{author}{{Morency}, L.}
  (\bibinfo{year}{2018}).
\newblock \bibinfo{title}{Openface 2.0: Facial behavior analysis toolkit}.
\newblock In {\it \bibinfo{booktitle}{AFGR}\/} (pp. \bibinfo{pages}{59--66}).
\newblock \DOIprefix\doi{10.1109/FG.2018.00019}.
%Type = Inproceedings
\bibitem[{{Browatzki} \& {Wallraven}(2020)}]{3FabRec}
\bibinfo{author}{{Browatzki}, B.}, \& \bibinfo{author}{{Wallraven}, C.}
  (\bibinfo{year}{2020}).
\newblock \bibinfo{title}{3fabrec: Fast few-shot face alignment by
  reconstruction}.
\newblock In {\it \bibinfo{booktitle}{CVPR}\/} (pp.
  \bibinfo{pages}{6109--6119}).
\newblock \DOIprefix\doi{10.1109/CVPR42600.2020.00615}.
%Type = Inproceedings
\bibitem[{{Bulat} \& {Tzimiropoulos}(2017)}]{MVB}
\bibinfo{author}{{Bulat}, A.}, \& \bibinfo{author}{{Tzimiropoulos}, G.}
  (\bibinfo{year}{2017}).
\newblock \bibinfo{title}{Binarized convolutional landmark localizers for human
  pose estimation and face alignment with limited resources}.
\newblock In {\it \bibinfo{booktitle}{ICCV}\/} (pp.
  \bibinfo{pages}{3726--3734}).
\newblock \DOIprefix\doi{10.1109/ICCV.2017.400}.
%Type = Inproceedings
\bibitem[{{Cao} et~al.(2012){Cao}, {Wei}, {Wen} \& {Sun}}]{ESR}
\bibinfo{author}{{Cao}, X.}, \bibinfo{author}{{Wei}, Y.},
  \bibinfo{author}{{Wen}, F.}, \& \bibinfo{author}{{Sun}, J.}
  (\bibinfo{year}{2012}).
\newblock \bibinfo{title}{Face alignment by explicit shape regression}.
\newblock In {\it \bibinfo{booktitle}{CVPR}\/} (pp.
  \bibinfo{pages}{2887--2894}).
\newblock \DOIprefix\doi{10.1109/CVPR.2012.6248015}.
%Type = Inproceedings
\bibitem[{{Chandran} et~al.(2020){Chandran}, {Bradley}, {Gross} \&
  {Beeler}}]{ADC}
\bibinfo{author}{{Chandran}, P.}, \bibinfo{author}{{Bradley}, D.},
  \bibinfo{author}{{Gross}, M.}, \& \bibinfo{author}{{Beeler}, T.}
  (\bibinfo{year}{2020}).
\newblock \bibinfo{title}{Attention-driven cropping for very high resolution
  facial landmark detection}.
\newblock In {\it \bibinfo{booktitle}{2020 IEEE/CVPR}\/} (pp.
  \bibinfo{pages}{5860--5869}).
\newblock \DOIprefix\doi{10.1109/CVPR42600.2020.00590}.
%Type = Inproceedings
\bibitem[{{Dapogny} et~al.(2019){Dapogny}, {Cord} \& {Bailly}}]{DeCaFA}
\bibinfo{author}{{Dapogny}, A.}, \bibinfo{author}{{Cord}, M.}, \&
  \bibinfo{author}{{Bailly}, K.} (\bibinfo{year}{2019}).
\newblock \bibinfo{title}{Decafa: Deep convolutional cascade for face alignment
  in the wild}.
\newblock In {\it \bibinfo{booktitle}{ICCV}\/} (pp.
  \bibinfo{pages}{6892--6900}).
\newblock \DOIprefix\doi{10.1109/ICCV.2019.00699}.
%Type = Inproceedings
\bibitem[{{Dong} et~al.(2018){Dong}, {Yan}, {Ouyang} \& {Yang}}]{SAN}
\bibinfo{author}{{Dong}, X.}, \bibinfo{author}{{Yan}, Y.},
  \bibinfo{author}{{Ouyang}, W.}, \& \bibinfo{author}{{Yang}, Y.}
  (\bibinfo{year}{2018}).
\newblock \bibinfo{title}{Style aggregated network for facial landmark
  detection}.
\newblock In {\it \bibinfo{booktitle}{CVPR}\/} (pp. \bibinfo{pages}{379--388}).
\newblock \DOIprefix\doi{10.1109/CVPR.2018.00047}.
%Type = Article
\bibitem[{Dong et~al.(2021)Dong, Yang, Wei, Weng, Sheikh \& Yu}]{SBR}
\bibinfo{author}{Dong, X.}, \bibinfo{author}{Yang, Y.}, \bibinfo{author}{Wei,
  S.-E.}, \bibinfo{author}{Weng, X.}, \bibinfo{author}{Sheikh, Y.}, \&
  \bibinfo{author}{Yu, S.-I.} (\bibinfo{year}{2021}).
\newblock \bibinfo{title}{Supervision by registration and triangulation for
  landmark detection}.
\newblock {\it \bibinfo{journal}{IEEE Transactions on Pattern Analysis and
  Machine Intelligence}\/},  {\it \bibinfo{volume}{43}\/},
  \bibinfo{pages}{3681--3694}. \DOIprefix\doi{10.1109/TPAMI.2020.2983935}.
%Type = Inproceedings
\bibitem[{{Dou} et~al.(2017){Dou}, {Shah} \& {Kakadiaris}}]{face_recostruction}
\bibinfo{author}{{Dou}, P.}, \bibinfo{author}{{Shah}, S.~K.}, \&
  \bibinfo{author}{{Kakadiaris}, I.~A.} (\bibinfo{year}{2017}).
\newblock \bibinfo{title}{End-to-end 3d face reconstruction with deep neural
  networks}.
\newblock In {\it \bibinfo{booktitle}{CVPR}\/} (pp.
  \bibinfo{pages}{1503--1512}).
\newblock \DOIprefix\doi{10.1109/CVPR.2017.164}.
%Type = Article
\bibitem[{D.{P. Kingma} \& J.{Ba}(2014)}]{Adam}
\bibinfo{author}{D.{P. Kingma}}, \& \bibinfo{author}{J.{Ba}}
  (\bibinfo{year}{2014}).
\newblock \bibinfo{title}{Adam: A method for stochastic optimization}.
\newblock {\it \bibinfo{journal}{arXiv preprint arXiv:1412.6980}\/}, .
%Type = Article
\bibitem[{{Drouard} et~al.(2017){Drouard}, {Horaud}, {Deleforge}, {Ba} \&
  {Evangelidis}}]{Robust_pose}
\bibinfo{author}{{Drouard}, V.}, \bibinfo{author}{{Horaud}, R.},
  \bibinfo{author}{{Deleforge}, A.}, \bibinfo{author}{{Ba}, S.}, \&
  \bibinfo{author}{{Evangelidis}, G.} (\bibinfo{year}{2017}).
\newblock \bibinfo{title}{Robust head-pose estimation based on partially-latent
  mixture of linear regressions}.
\newblock {\it \bibinfo{journal}{IEEE Transactions on Image Processing}\/},
  {\it \bibinfo{volume}{26}\/}, \bibinfo{pages}{1428--1440}.
  \DOIprefix\doi{10.1109/TIP.2017.2654165}.
%Type = Article
\bibitem[{Esen et~al.(2017)Esen, Esen \& Ozsolak}]{MEP}
\bibinfo{author}{Esen, H.}, \bibinfo{author}{Esen, M.}, \&
  \bibinfo{author}{Ozsolak, O.} (\bibinfo{year}{2017}).
\newblock \bibinfo{title}{Modelling and experimental performance analysis of
  solar-assisted ground source heat pump system}.
\newblock {\it \bibinfo{journal}{Journal of Experimental \& Theoretical
  Artificial Intelligence}\/},  {\it \bibinfo{volume}{29}\/},
  \bibinfo{pages}{1--17}. \DOIprefix\doi{10.1080/0952813X.2015.1056242}.
%Type = Article
\bibitem[{Esen et~al.(2008{\natexlab{a}})Esen, Inalli, Sengur \& Esen}]{ANNA}
\bibinfo{author}{Esen, H.}, \bibinfo{author}{Inalli, M.},
  \bibinfo{author}{Sengur, A.}, \& \bibinfo{author}{Esen, M.}
  (\bibinfo{year}{2008}{\natexlab{a}}).
\newblock \bibinfo{title}{Artificial neural networks and adaptive neuro-fuzzy
  assessments for ground-coupled heat pump system}.
\newblock {\it \bibinfo{journal}{Energy and Buildings}\/},  {\it
  \bibinfo{volume}{40}\/}, \bibinfo{pages}{1074--1083}.
  \DOIprefix\doi{https://doi.org/10.1016/j.enbuild.2007.10.002}.
%Type = Article
\bibitem[{Esen et~al.(2008{\natexlab{b}})Esen, Inalli, Sengur \& Esen}]{FOG}
\bibinfo{author}{Esen, H.}, \bibinfo{author}{Inalli, M.},
  \bibinfo{author}{Sengur, A.}, \& \bibinfo{author}{Esen, M.}
  (\bibinfo{year}{2008}{\natexlab{b}}).
\newblock \bibinfo{title}{Forecasting of a ground-coupled heat pump performance
  using neural networks with statistical data weighting pre-processing}.
\newblock {\it \bibinfo{journal}{International Journal of Thermal Sciences}\/},
   {\it \bibinfo{volume}{47}\/}, \bibinfo{pages}{431--441}.
  \DOIprefix\doi{https://doi.org/10.1016/j.ijthermalsci.2007.03.004}.
%Type = Article
\bibitem[{Esen et~al.(2008{\natexlab{c}})Esen, Inalli, Sengur \& Esen}]{PPO}
\bibinfo{author}{Esen, H.}, \bibinfo{author}{Inalli, M.},
  \bibinfo{author}{Sengur, A.}, \& \bibinfo{author}{Esen, M.}
  (\bibinfo{year}{2008}{\natexlab{c}}).
\newblock \bibinfo{title}{Performance prediction of a ground-coupled heat pump
  system using artificial neural networks}.
\newblock {\it \bibinfo{journal}{Expert Systems with Applications}\/},  {\it
  \bibinfo{volume}{35}\/}, \bibinfo{pages}{1940--1948}.
  \DOIprefix\doi{https://doi.org/10.1016/j.eswa.2007.08.081}.
%Type = Article
\bibitem[{Esen et~al.(2009)Esen, Ozgen, Esen \& Sengur}]{ANNW}
\bibinfo{author}{Esen, H.}, \bibinfo{author}{Ozgen, F.}, \bibinfo{author}{Esen,
  M.}, \& \bibinfo{author}{Sengur, A.} (\bibinfo{year}{2009}).
\newblock \bibinfo{title}{Artificial neural network and wavelet neural network
  approaches for modelling of a solar air heater}.
\newblock {\it \bibinfo{journal}{Expert Systems with Applications}\/},  {\it
  \bibinfo{volume}{36}\/}, \bibinfo{pages}{11240--11248}.
  \DOIprefix\doi{https://doi.org/10.1016/j.eswa.2009.02.073}.
%Type = Inproceedings
\bibitem[{{Hassner} et~al.(2015){Hassner}, {Harel}, {Paz} \&
  {Enbar}}]{face_morphing}
\bibinfo{author}{{Hassner}, T.}, \bibinfo{author}{{Harel}, S.},
  \bibinfo{author}{{Paz}, E.}, \& \bibinfo{author}{{Enbar}, R.}
  (\bibinfo{year}{2015}).
\newblock \bibinfo{title}{Effective face frontalization in unconstrained
  images}.
\newblock In {\it \bibinfo{booktitle}{CVPR}\/} (pp.
  \bibinfo{pages}{4295--4304}).
\newblock \DOIprefix\doi{10.1109/CVPR.2015.7299058}.
%Type = Inproceedings
\bibitem[{{Howard} et~al.(2019){Howard}, {Sandler}, {Chen}, {Wang}, {Chen},
  {Tan}, {Chu}, {Vasudevan}, {Zhu}, {Pang}, {Adam} \& {Le}}]{mobileV3}
\bibinfo{author}{{Howard}, A.}, \bibinfo{author}{{Sandler}, M.},
  \bibinfo{author}{{Chen}, B.}, \bibinfo{author}{{Wang}, W.},
  \bibinfo{author}{{Chen}, L.}, \bibinfo{author}{{Tan}, M.},
  \bibinfo{author}{{Chu}, G.}, \bibinfo{author}{{Vasudevan}, V.},
  \bibinfo{author}{{Zhu}, Y.}, \bibinfo{author}{{Pang}, R.},
  \bibinfo{author}{{Adam}, H.}, \& \bibinfo{author}{{Le}, Q.}
  (\bibinfo{year}{2019}).
\newblock \bibinfo{title}{Searching for mobilenetv3}.
\newblock In {\it \bibinfo{booktitle}{ICCV}\/} (pp.
  \bibinfo{pages}{1314--1324}).
\newblock \DOIprefix\doi{10.1109/ICCV.2019.00140}.
%Type = Article
\bibitem[{{Huang} et~al.(1995){Huang}, {Bruckstein}, {Holt} \&
  {Netravali}}]{U3D-pose}
\bibinfo{author}{{Huang}, T.~S.}, \bibinfo{author}{{Bruckstein}, A.~M.},
  \bibinfo{author}{{Holt}, R.~J.}, \& \bibinfo{author}{{Netravali}, A.~N.}
  (\bibinfo{year}{1995}).
\newblock \bibinfo{title}{Uniqueness of 3d pose under weak perspective: a
  geometrical proof}.
\newblock {\it \bibinfo{journal}{IEEE Transactions on Pattern Analysis and
  Machine Intelligence}\/},  {\it \bibinfo{volume}{17}\/},
  \bibinfo{pages}{1220--1221}. \DOIprefix\doi{10.1109/34.476515}.
%Type = Inproceedings
\bibitem[{{Huang} et~al.(2020){Huang}, {Deng}, {Shen}, {Zhang} \&
  {Ye}}]{PropNet}
\bibinfo{author}{{Huang}, X.}, \bibinfo{author}{{Deng}, W.},
  \bibinfo{author}{{Shen}, H.}, \bibinfo{author}{{Zhang}, X.}, \&
  \bibinfo{author}{{Ye}, J.} (\bibinfo{year}{2020}).
\newblock \bibinfo{title}{Propagationnet: Propagate points to curve to learn
  structure information}.
\newblock In {\it \bibinfo{booktitle}{CVPR}\/} (pp.
  \bibinfo{pages}{7263--7272}).
\newblock \DOIprefix\doi{10.1109/CVPR42600.2020.00729}.
%Type = Inproceedings
\bibitem[{Lin et~al.(2017)Lin, Dollár, Girshick, He, Hariharan \&
  Belongie}]{FPN}
\bibinfo{author}{Lin, T.-Y.}, \bibinfo{author}{Dollár, P.},
  \bibinfo{author}{Girshick, R.}, \bibinfo{author}{He, K.},
  \bibinfo{author}{Hariharan, B.}, \& \bibinfo{author}{Belongie, S.}
  (\bibinfo{year}{2017}).
\newblock \bibinfo{title}{Feature pyramid networks for object detection}.
\newblock In {\it \bibinfo{booktitle}{CVPR}\/} (pp. \bibinfo{pages}{936--944}).
\newblock \DOIprefix\doi{10.1109/CVPR.2017.106}.
%Type = Article
\bibitem[{{Liu} et~al.(2018){Liu}, {Lu}, {Feng} \& {Zhou}}]{Two-stream_video}
\bibinfo{author}{{Liu}, H.}, \bibinfo{author}{{Lu}, J.},
  \bibinfo{author}{{Feng}, J.}, \& \bibinfo{author}{{Zhou}, J.}
  (\bibinfo{year}{2018}).
\newblock \bibinfo{title}{Two-stream transformer networks for video-based face
  alignment}.
\newblock {\it \bibinfo{journal}{IEEE Transactions on Pattern Analysis and
  Machine Intelligence}\/},  {\it \bibinfo{volume}{40}\/},
  \bibinfo{pages}{2546--2554}. \DOIprefix\doi{10.1109/TPAMI.2017.2734779}.
%Type = Article
\bibitem[{{Liu} et~al.(2019){Liu}, {Wang}, {Zhu}, {Mo}, {Wang}, {Yin}, {Shi} \&
  {Wei}}]{ACFSS}
\bibinfo{author}{{Liu}, L.}, \bibinfo{author}{{Wang}, Q.},
  \bibinfo{author}{{Zhu}, W.}, \bibinfo{author}{{Mo}, H.},
  \bibinfo{author}{{Wang}, T.}, \bibinfo{author}{{Yin}, S.},
  \bibinfo{author}{{Shi}, Y.}, \& \bibinfo{author}{{Wei}, S.}
  (\bibinfo{year}{2019}).
\newblock \bibinfo{title}{A face alignment accelerator based on optimized
  coarse-to-fine shape searching}.
\newblock {\it \bibinfo{journal}{IEEE Transactions on Circuits and Systems for
  Video Technology}\/},  {\it \bibinfo{volume}{29}\/},
  \bibinfo{pages}{2467--2481}. \DOIprefix\doi{10.1109/TCSVT.2018.2867499}.
%Type = Inproceedings
\bibitem[{Liu et~al.(2020)Liu, Shi, Shen, Si, Wang \& Mei}]{LAPA}
\bibinfo{author}{Liu, Y.}, \bibinfo{author}{Shi, H.}, \bibinfo{author}{Shen,
  H.}, \bibinfo{author}{Si, Y.}, \bibinfo{author}{Wang, X.}, \&
  \bibinfo{author}{Mei, T.} (\bibinfo{year}{2020}).
\newblock \bibinfo{title}{A new dataset and boundary-attention semantic
  segmentation for face parsing}.
\newblock In {\it \bibinfo{booktitle}{AAAI 2020}\/} (pp.
  \bibinfo{pages}{11637--11644}).
\newblock \URLprefix
  \url{https://aaai.org/ojs/index.php/AAAI/article/view/6832}.
%Type = Inproceedings
\bibitem[{{Martins} \& {Batista}(2008)}]{accurate_model}
\bibinfo{author}{{Martins}, P.}, \& \bibinfo{author}{{Batista}, J.}
  (\bibinfo{year}{2008}).
\newblock \bibinfo{title}{Accurate single view model-based head pose
  estimation}.
\newblock In {\it \bibinfo{booktitle}{AFGR}\/} (pp. \bibinfo{pages}{1--6}).
\newblock \DOIprefix\doi{10.1109/AFGR.2008.4813369}.
%Type = Article
\bibitem[{{Mo} et~al.(2020){Mo}, {Liu}, {Zhu}, {Li}, {Liu}, {Yin} \&
  {Wei}}]{Hard-MTCNN}
\bibinfo{author}{{Mo}, H.}, \bibinfo{author}{{Liu}, L.},
  \bibinfo{author}{{Zhu}, W.}, \bibinfo{author}{{Li}, Q.},
  \bibinfo{author}{{Liu}, H.}, \bibinfo{author}{{Yin}, S.}, \&
  \bibinfo{author}{{Wei}, S.} (\bibinfo{year}{2020}).
\newblock \bibinfo{title}{A multi-task hardwired accelerator for face detection
  and alignment}.
\newblock {\it \bibinfo{journal}{IEEE Transactions on Circuits and Systems for
  Video Technology}\/},  {\it \bibinfo{volume}{30}\/},
  \bibinfo{pages}{4284--4298}. \DOIprefix\doi{10.1109/TCSVT.2019.2955463}.
%Type = Inproceedings
\bibitem[{Newell et~al.(2016)Newell, Yang \& Deng}]{Hourglass}
\bibinfo{author}{Newell, A.}, \bibinfo{author}{Yang, K.}, \&
  \bibinfo{author}{Deng, J.} (\bibinfo{year}{2016}).
\newblock \bibinfo{title}{Stacked hourglass networks for human pose
  estimation}.
\newblock In \bibinfo{editor}{B.~Leibe}, \bibinfo{editor}{J.~Matas},
  \bibinfo{editor}{N.~Sebe}, \& \bibinfo{editor}{M.~Welling} (Eds.), {\it
  \bibinfo{booktitle}{Computer Vision -- ECCV 2016}\/} (pp.
  \bibinfo{pages}{483--499}).
\newblock \bibinfo{address}{Cham}: \bibinfo{publisher}{Springer International
  Publishing}.
%Type = Article
\bibitem[{Patacchiola \& Cangelosi(2017)}]{CNN-Pose}
\bibinfo{author}{Patacchiola, M.}, \& \bibinfo{author}{Cangelosi, A.}
  (\bibinfo{year}{2017}).
\newblock \bibinfo{title}{Head pose estimation in the wild using convolutional
  neural networks and adaptive gradient methods}.
\newblock {\it \bibinfo{journal}{Pattern Recognition}\/},  {\it
  \bibinfo{volume}{71}\/}, \bibinfo{pages}{132 -- 143}.
  \DOIprefix\doi{https://doi.org/10.1016/j.patcog.2017.06.009}.
%Type = Inproceedings
\bibitem[{{Qian} et~al.(2019){Qian}, {Sun}, {Wu}, {Qian} \& {Jia}}]{AVS}
\bibinfo{author}{{Qian}, S.}, \bibinfo{author}{{Sun}, K.},
  \bibinfo{author}{{Wu}, W.}, \bibinfo{author}{{Qian}, C.}, \&
  \bibinfo{author}{{Jia}, J.} (\bibinfo{year}{2019}).
\newblock \bibinfo{title}{Aggregation via separation: Boosting facial landmark
  detector with semi-supervised style translation}.
\newblock In {\it \bibinfo{booktitle}{ICCV}\/} (pp.
  \bibinfo{pages}{10152--10162}).
\newblock \DOIprefix\doi{10.1109/ICCV.2019.01025}.
%Type = Article
\bibitem[{{Ranjan} et~al.(2019){Ranjan}, {Patel} \& {Chellappa}}]{Hyperface}
\bibinfo{author}{{Ranjan}, R.}, \bibinfo{author}{{Patel}, V.~M.}, \&
  \bibinfo{author}{{Chellappa}, R.} (\bibinfo{year}{2019}).
\newblock \bibinfo{title}{Hyperface: A deep multi-task learning framework for
  face detection, landmark localization, pose estimation, and gender
  recognition}.
\newblock {\it \bibinfo{journal}{IEEE Transactions on Pattern Analysis and
  Machine Intelligence}\/},  {\it \bibinfo{volume}{41}\/},
  \bibinfo{pages}{121--135}. \DOIprefix\doi{10.1109/TPAMI.2017.2781233}.
%Type = Inproceedings
\bibitem[{R.{Liu} et~al.(2018)R.{Liu}, L.{Joel}, P.{Molino}, F.P.{Such},
  E.{Frank}, A.{Sergeev} \& J.{Yosinski}}]{CoordCNN}
\bibinfo{author}{R.{Liu}}, \bibinfo{author}{L.{Joel}},
  \bibinfo{author}{P.{Molino}}, \bibinfo{author}{F.P.{Such}},
  \bibinfo{author}{E.{Frank}}, \bibinfo{author}{A.{Sergeev}}, \&
  \bibinfo{author}{J.{Yosinski}} (\bibinfo{year}{2018}).
\newblock \bibinfo{title}{An intriguing failing of convolutional neural
  networks and the coordconv solution}.
\newblock In {\it \bibinfo{booktitle}{NIPS}\/} NIPS'18 (p.
  \bibinfo{pages}{9628–9639}).
\newblock \bibinfo{address}{Red Hook, NY, USA}: \bibinfo{publisher}{Curran
  Associates Inc.}
%Type = Inproceedings
\bibitem[{{Ruiz} et~al.(2018){Ruiz}, {Chong} \& {Rehg}}]{Ruzi}
\bibinfo{author}{{Ruiz}, N.}, \bibinfo{author}{{Chong}, E.}, \&
  \bibinfo{author}{{Rehg}, J.~M.} (\bibinfo{year}{2018}).
\newblock \bibinfo{title}{Fine-grained head pose estimation without keypoints}.
\newblock In {\it \bibinfo{booktitle}{2018 IEEE/CVF Conference on Computer
  Vision and Pattern Recognition Workshops (CVPRW)}\/} (pp.
  \bibinfo{pages}{2155--215509}).
\newblock \DOIprefix\doi{10.1109/CVPRW.2018.00281}.
%Type = Inproceedings
\bibitem[{{Sagonas} et~al.(2013){Sagonas}, {Tzimiropoulos}, {Zafeiriou} \&
  {Pantic}}]{300W}
\bibinfo{author}{{Sagonas}, C.}, \bibinfo{author}{{Tzimiropoulos}, G.},
  \bibinfo{author}{{Zafeiriou}, S.}, \& \bibinfo{author}{{Pantic}, M.}
  (\bibinfo{year}{2013}).
\newblock \bibinfo{title}{300 faces in-the-wild challenge: The first facial
  landmark localization challenge}.
\newblock In {\it \bibinfo{booktitle}{ICCVW}\/} (pp.
  \bibinfo{pages}{397--403}).
\newblock \DOIprefix\doi{10.1109/ICCVW.2013.59}.
%Type = Article
\bibitem[{Shao et~al.(2021)Shao, Xing, Lyu, Zhou, Shi \& Maybank}]{RFA}
\bibinfo{author}{Shao, X.}, \bibinfo{author}{Xing, J.}, \bibinfo{author}{Lyu,
  J.}, \bibinfo{author}{Zhou, X.}, \bibinfo{author}{Shi, Y.}, \&
  \bibinfo{author}{Maybank, S.~J.} (\bibinfo{year}{2021}).
\newblock \bibinfo{title}{Robust face alignment via deep progressive
  reinitialization and adaptive error-driven learning}.
\newblock {\it \bibinfo{journal}{IEEE Transactions on Pattern Analysis and
  Machine Intelligence}\/},  (pp. \bibinfo{pages}{1--1}).
  \DOIprefix\doi{10.1109/TPAMI.2021.3073593}.
%Type = Inproceedings
\bibitem[{Shapira et~al.(2021)Shapira, Levy, Goldin \& Jevnisek}]{KWTQ}
\bibinfo{author}{Shapira, G.}, \bibinfo{author}{Levy, N.},
  \bibinfo{author}{Goldin, I.}, \& \bibinfo{author}{Jevnisek, R.~J.}
  (\bibinfo{year}{2021}).
\newblock \bibinfo{title}{Knowing when to quit: Selective cascaded regression
  with patch attention for real-time face alignment}.
\newblock In {\it \bibinfo{booktitle}{Proceedings of the 29th ACM International
  Conference on Multimedia}\/} (pp. \bibinfo{pages}{2372--2380}).
%Type = Inproceedings
\bibitem[{{Shen} et~al.(2015){Shen}, {Zafeiriou}, {Chrysos}, {Kossaifi},
  {Tzimiropoulos} \& {Pantic}}]{300VW}
\bibinfo{author}{{Shen}, J.}, \bibinfo{author}{{Zafeiriou}, S.},
  \bibinfo{author}{{Chrysos}, G.~G.}, \bibinfo{author}{{Kossaifi}, J.},
  \bibinfo{author}{{Tzimiropoulos}, G.}, \& \bibinfo{author}{{Pantic}, M.}
  (\bibinfo{year}{2015}).
\newblock \bibinfo{title}{The first facial landmark tracking in-the-wild
  challenge: Benchmark and results}.
\newblock In {\it \bibinfo{booktitle}{ICCVW}\/} (pp.
  \bibinfo{pages}{1003--1011}).
\newblock \DOIprefix\doi{10.1109/ICCVW.2015.132}.
%Type = Inproceedings
\bibitem[{Simonyan \& Zisserman(2014)}]{Two-Stream-action}
\bibinfo{author}{Simonyan, K.}, \& \bibinfo{author}{Zisserman, A.}
  (\bibinfo{year}{2014}).
\newblock \bibinfo{title}{Two-stream convolutional networks for action
  recognition in videos}.
\newblock In {\it \bibinfo{booktitle}{NIPS}\/} NIPS'14 (p.
  \bibinfo{pages}{568–576}).
\newblock \bibinfo{address}{Cambridge, MA, USA}: \bibinfo{publisher}{MIT
  Press}.
%Type = Inproceedings
\bibitem[{{Sun} et~al.(2019){Sun}, {Wu}, {Liu}, {Yang}, {Wang}, {Zhou}, {Ye} \&
  {Qian}}]{FAB}
\bibinfo{author}{{Sun}, K.}, \bibinfo{author}{{Wu}, W.},
  \bibinfo{author}{{Liu}, T.}, \bibinfo{author}{{Yang}, S.},
  \bibinfo{author}{{Wang}, Q.}, \bibinfo{author}{{Zhou}, Q.},
  \bibinfo{author}{{Ye}, Z.}, \& \bibinfo{author}{{Qian}, C.}
  (\bibinfo{year}{2019}).
\newblock \bibinfo{title}{Fab: A robust facial landmark detection framework for
  motion-blurred videos}.
\newblock In {\it \bibinfo{booktitle}{ICCV}\/} (pp.
  \bibinfo{pages}{5461--5470}).
\newblock \DOIprefix\doi{10.1109/ICCV.2019.00556}.
%Type = Inproceedings
\bibitem[{{Wu} et~al.(2018{\natexlab{a}}){Wu}, {Qian}, {Yang}, {Wang}, {Cai} \&
  {Zhou}}]{LAB}
\bibinfo{author}{{Wu}, W.}, \bibinfo{author}{{Qian}, C.},
  \bibinfo{author}{{Yang}, S.}, \bibinfo{author}{{Wang}, Q.},
  \bibinfo{author}{{Cai}, Y.}, \& \bibinfo{author}{{Zhou}, Q.}
  (\bibinfo{year}{2018}{\natexlab{a}}).
\newblock \bibinfo{title}{Look at boundary: A boundary-aware face alignment
  algorithm}.
\newblock In {\it \bibinfo{booktitle}{CVPR}\/} (pp.
  \bibinfo{pages}{2129--2138}).
\newblock \DOIprefix\doi{10.1109/CVPR.2018.00227}.
%Type = Inproceedings
\bibitem[{{Wu} \& {Yang}(2017)}]{DVLN}
\bibinfo{author}{{Wu}, W.}, \& \bibinfo{author}{{Yang}, S.}
  (\bibinfo{year}{2017}).
\newblock \bibinfo{title}{Leveraging intra and inter-dataset variations for
  robust face alignment}.
\newblock In {\it \bibinfo{booktitle}{CVPRW}\/} (pp.
  \bibinfo{pages}{2096--2105}).
\newblock \DOIprefix\doi{10.1109/CVPRW.2017.261}.
%Type = Inproceedings
\bibitem[{{Wu} et~al.(2018{\natexlab{b}}){Wu}, {Yin}, {Wang}, {Wang} \&
  {Xu}}]{facial_expression}
\bibinfo{author}{{Wu}, W.}, \bibinfo{author}{{Yin}, Y.},
  \bibinfo{author}{{Wang}, Y.}, \bibinfo{author}{{Wang}, X.}, \&
  \bibinfo{author}{{Xu}, D.} (\bibinfo{year}{2018}{\natexlab{b}}).
\newblock \bibinfo{title}{Facial expression recognition for different pose
  faces based on special landmark detection}.
\newblock In {\it \bibinfo{booktitle}{ICPR}\/} (pp.
  \bibinfo{pages}{1524--1529}).
\newblock \DOIprefix\doi{10.1109/ICPR.2018.8545725}.
%Type = Inproceedings
\bibitem[{{Xiong} \& {De la Torre}(2013)}]{SDM}
\bibinfo{author}{{Xiong}, X.}, \& \bibinfo{author}{{De la Torre}, F.}
  (\bibinfo{year}{2013}).
\newblock \bibinfo{title}{Supervised descent method and its applications to
  face alignment}.
\newblock In {\it \bibinfo{booktitle}{CVPR}\/} (pp. \bibinfo{pages}{532--539}).
\newblock \DOIprefix\doi{10.1109/CVPR.2013.75}.
%Type = Inproceedings
\bibitem[{{Yang} et~al.(2016){Yang}, {Luo}, {Loy} \& {Tang}}]{WIDER}
\bibinfo{author}{{Yang}, S.}, \bibinfo{author}{{Luo}, P.},
  \bibinfo{author}{{Loy}, C.~C.}, \& \bibinfo{author}{{Tang}, X.}
  (\bibinfo{year}{2016}).
\newblock \bibinfo{title}{Wider face: A face detection benchmark}.
\newblock In {\it \bibinfo{booktitle}{CVPR}\/} (pp.
  \bibinfo{pages}{5525--5533}).
\newblock \DOIprefix\doi{10.1109/CVPR.2016.596}.
%Type = Inproceedings
\bibitem[{Yang et~al.(2019)Yang, Chen, Lin \& Chuang}]{FSA-Net}
\bibinfo{author}{Yang, T.-Y.}, \bibinfo{author}{Chen, Y.-T.},
  \bibinfo{author}{Lin, Y.-Y.}, \& \bibinfo{author}{Chuang, Y.-Y.}
  (\bibinfo{year}{2019}).
\newblock \bibinfo{title}{Fsa-net: Learning fine-grained structure aggregation
  for head pose estimation from a single image}.
\newblock In {\it \bibinfo{booktitle}{2019 IEEE/CVF Conference on Computer
  Vision and Pattern Recognition (CVPR)}\/} (pp. \bibinfo{pages}{1087--1096}).
\newblock \DOIprefix\doi{10.1109/CVPR.2019.00118}.
%Type = Inproceedings
\bibitem[{{Zadeh} et~al.(2017){Zadeh}, {Baltrušaitis} \& {Morency}}]{CE-CLM}
\bibinfo{author}{{Zadeh}, A.}, \bibinfo{author}{{Baltrušaitis}, T.}, \&
  \bibinfo{author}{{Morency}, L.} (\bibinfo{year}{2017}).
\newblock \bibinfo{title}{Convolutional experts constrained local model for
  facial landmark detection}.
\newblock In {\it \bibinfo{booktitle}{CVPRW}\/} (pp.
  \bibinfo{pages}{2051--2059}).
\newblock \DOIprefix\doi{10.1109/CVPRW.2017.256}.
%Type = Inproceedings
\bibitem[{{Zhang} \& {Zhang}(2014)}]{zhang-multi}
\bibinfo{author}{{Zhang}, C.}, \& \bibinfo{author}{{Zhang}, Z.}
  (\bibinfo{year}{2014}).
\newblock \bibinfo{title}{Improving multiview face detection with multi-task
  deep convolutional neural networks}.
\newblock In {\it \bibinfo{booktitle}{WACV}\/} (pp.
  \bibinfo{pages}{1036--1041}).
\newblock \DOIprefix\doi{10.1109/WACV.2014.6835990}.
%Type = Article
\bibitem[{{Zhang} et~al.(2016){Zhang}, {Zhang}, {Li} \& {Qiao}}]{MTCNN}
\bibinfo{author}{{Zhang}, K.}, \bibinfo{author}{{Zhang}, Z.},
  \bibinfo{author}{{Li}, Z.}, \& \bibinfo{author}{{Qiao}, Y.}
  (\bibinfo{year}{2016}).
\newblock \bibinfo{title}{Joint face detection and alignment using multitask
  cascaded convolutional networks}.
\newblock {\it \bibinfo{journal}{IEEE Signal Processing Letters}\/},  {\it
  \bibinfo{volume}{23}\/}, \bibinfo{pages}{1499--1503}.
  \DOIprefix\doi{10.1109/LSP.2016.2603342}.
%Type = Inproceedings
\bibitem[{Zhu et~al.(2015)Zhu, Li, Loy \& Tang}]{CFSS}
\bibinfo{author}{Zhu, S.}, \bibinfo{author}{Li, C.}, \bibinfo{author}{Loy,
  C.~C.}, \& \bibinfo{author}{Tang, X.} (\bibinfo{year}{2015}).
\newblock \bibinfo{title}{Face alignment by coarse-to-fine shape searching}.
\newblock In {\it \bibinfo{booktitle}{CVPR}\/} (pp.
  \bibinfo{pages}{4998--5006}).
\newblock \DOIprefix\doi{10.1109/CVPR.2015.7299134}.
%Type = Inproceedings
\bibitem[{{Zhu} et~al.(2016){Zhu}, {Lei}, {Liu}, {Shi} \& {Li}}]{3DDFA}
\bibinfo{author}{{Zhu}, X.}, \bibinfo{author}{{Lei}, Z.},
  \bibinfo{author}{{Liu}, X.}, \bibinfo{author}{{Shi}, H.}, \&
  \bibinfo{author}{{Li}, S.~Z.} (\bibinfo{year}{2016}).
\newblock \bibinfo{title}{Face alignment across large poses: A 3d solution}.
\newblock In {\it \bibinfo{booktitle}{CVPR}\/} (pp. \bibinfo{pages}{146--155}).
\newblock \DOIprefix\doi{10.1109/CVPR.2016.23}.
%Type = Article
\bibitem[{{Zou} et~al.(2020){Zou}, {Xiao}, {Wang}, {Yan}, {Zhong} \&
  {Wu}}]{TUF}
\bibinfo{author}{{Zou}, X.}, \bibinfo{author}{{Xiao}, P.},
  \bibinfo{author}{{Wang}, J.}, \bibinfo{author}{{Yan}, L.},
  \bibinfo{author}{{Zhong}, S.}, \& \bibinfo{author}{{Wu}, Y.}
  (\bibinfo{year}{2020}).
\newblock \bibinfo{title}{Towards unconstrained facial landmark detection
  robust to diverse cropping manners}.
\newblock {\it \bibinfo{journal}{IEEE Transactions on Circuits and Systems for
  Video Technology}\/},  (pp. \bibinfo{pages}{1--1}).
  \DOIprefix\doi{10.1109/TCSVT.2020.3006236}.

\end{thebibliography}

\end{document}